\documentclass{article}

\PassOptionsToPackage{numbers,sort&compress}{natbib}

\usepackage[main, final]{neurips_2026}
\usepackage{enumitem}
\usepackage[utf8]{inputenc}
\usepackage[T1]{fontenc}
\usepackage{hyperref}
\usepackage{url}
\usepackage{booktabs}
\usepackage{graphicx}
\usepackage{amsfonts}
\usepackage{nicefrac}
\usepackage{microtype}
\usepackage{xcolor}
\usepackage{amsmath,amssymb}
\usepackage{tikz}
\usepackage{tikz-cd}
\usepackage{multicol}
\usepackage{algorithm}
\usepackage[noend]{algpseudocode}
\usetikzlibrary{arrows.meta,positioning,shapes.geometric}
\graphicspath{{Figures/}}
\hypersetup{hypertexnames=false}
\algrenewcommand\algorithmicrequire{\textbf{Input:}}
\algrenewcommand\algorithmicensure{\textbf{Output:}}

\usepackage{amsmath,amssymb}
\usepackage{amsthm}

\newtheorem{theorem}{Theorem}
\newtheorem{lemma}[theorem]{Lemma}
\newtheorem{proposition}[theorem]{Proposition}

\newtheorem{definition}[theorem]{Definition}

\newcommand{\sizeboundedproblemtitle}{MSE-$k$-NAM}
\newcommand{\sizeboundedproblemname}{\hyperlink{prob:any-order-size-bounded}{\sizeboundedproblemtitle\ }}

\title{Do Feature Interactions Break Explainability? \\ Provable Explanations for Any-Order Neural Additive Models}

\author{%
  Idan Refaeli\thanks{Corresponding author: \texttt{idan.refaeli@mail.huji.ac.il}.} \\
  The Hebrew University of Jerusalem
  \And
  Yizhak Yisrael Elboher \\
  The Hebrew University of Jerusalem\\ Masaryk University
  \And
  Guy Katz \\
  The Hebrew University of Jerusalem
  \And
  Shahaf Bassan \\
  The Hebrew University of Jerusalem
}

\begin{document}

\maketitle

\begin{abstract}
Post-hoc explanation methods for neural networks are often heuristic and lack
provable guarantees. A common alternative to such methods that was studied in recent years is to compute a
\emph{cardinally-minimal} subset of input features that is
\emph{provably sufficient} to determine the model's prediction.
For general neural networks, however, finding such explanations is computationally intractable. In contrast, recent work showed that this problem becomes more tractable for a particular family of neural networks, namely neural additive models (NAMs), though standard NAMs are highly restrictive because they exclude feature interactions altogether. In this work, we study the computation of provably
minimal and sufficient explanations for neural networks with restricted \emph{feature
interactions}, bridging the gap between fully general neural networks and
purely additive models. We first prove that even with constant-order
interactions, the problem is $\Sigma_2^P$-hard, implying that worst-case
exponential complexity is unavoidable. We then identify two provably tractable
settings:
one based on sparsity in the interaction structure and another based on a
stronger interaction-aware notion of sufficiency. Under either of these settings, we
develop algorithms that compute provable explanations for neural networks with
any-order interactions while retaining much of the computational efficiency of
additive models. Empirically, we show that these explanations are significantly smaller and
faster to obtain than those produced by standard algorithms, while being
derived from models that achieve
much higher accuracy than standard NAMs. Overall, our results help
expand the class of neural networks that admit efficient provable explanations,
while clarifying the fundamental role of feature interactions in shaping their
complexity.
\end{abstract}

\section{Introduction}

Understanding why neural networks make particular predictions remains one of
the central challenges in trustworthy ML. A significant amount of prior work
has proposed \emph{post-hoc explanation} methods, aiming to identify which
input features are most responsible for a model’s decision. Popular approaches
such as LIME~\cite{RiSiGu16}, SHAP~\cite{LuLe17}, and Integrated Gradients~\cite{SuTaYa17} assign attribution scores to
features, while subset-selection based methods such as Anchors~\cite{RiSiGu18} and SIS~\cite{CaMuJaGi19} seek small subsets of observed features that suffice to justify
a prediction. These methods have become standard tools in practice, but they
remain largely heuristic: their outputs typically rely on local approximations,
sampling, or specific perturbation assumptions, and therefore do not provide
provable guarantees.

Recent work has increasingly focused on \emph{provable explanations}~\citep{WuWuBa23, IgNaMa19, BaElGu25, IzHuMoPlIgMa24, LaZbMiPaKw21, DaHi20, AuBeBoKoLaMa21}: explanations
equipped with formal certificates guaranteeing that the model’s prediction is
preserved under all admissible completions of the unobserved features. Among
these, \emph{provably sufficient explanations}~\citep{WuWuBa23, IgNaMa19, BaElGu25, IzHuMoPlIgMa24} provide a particularly strong
guarantee: fixing only the features in the explanation suffices to preserve the
prediction regardless of how the remaining features vary. However, sufficiency
alone is not enough: the full input trivially satisfies this property. For this
reason, the gold-standard objective is to compute \emph{cardinally-minimal sufficient
explanations}~\citep{WuWuBa23, IgNaMa19, BaElGu25, IzHuMoPlIgMa24}, which identify the smallest subset of features that provably
preserves the prediction. In practice, most formal explainability work instead
targets the weaker \emph{subset-minimal} notion, since cardinal minimality is
harder to compute; cardinal minimality still provides both formal correctness
and conciseness, making it one of the strongest explanation guarantees studied in
the literature.

Unfortunately, these guarantees come at a steep computational cost. For
general neural networks, computing cardinally-minimal provably sufficient
explanations is known to be $\Sigma_2^P$-hard~\citep{BaMoPeSu20}, placing the problem at the
second level of the polynomial hierarchy. Intuitively, this means that this problem is substantially harder than
NP-Complete problems: it requires dispatching, in the worst-case, an \emph{exponential} number of
neural-network verification queries, each of which is itself NP-hard. Thus,
while these provable explanations offer stronger guarantees than heuristic
alternatives, they become computationally intractable precisely for the models
where explanations are most needed.

A recent line of work provides an encouraging exception by focusing on a
specific family of neural networks: \emph{Neural Additive Models} (NAMs)~\citep{AgMeFrZhLeCaHi21}.
These networks decompose predictions into sums of feature-specific components,
and have received substantial attention in recent years~\citep{AgMeFrZhLeCaHi21, RaDuMa22, BeGlGi24, KiDiSa24, ZhBaPa24, EnLi25}. Recent work~\citep{BaElLaSaKrAlKa26} showed that cardinally-minimal provably sufficient explanations can be
computed in this setting using only a worst-case \emph{linear} number of verification
queries. This demonstrates that model structure can fundamentally change the
computational landscape of computing provable explanations. However, standard NAMs are highly
restrictive. By construction, they exclude feature interactions entirely,
forcing each feature to contribute independently to the prediction. While this
pure additivity enables tractability, it limits expressivity and can reduce
predictive performance relative to models that explicitly represent
interactions~\citep{LoCaGeHo13, CaLoGeKoStEl15, EnLi22}. In many real-world settings, predictions depend critically on
feature interactions, making this restriction a practical limitation.

\textbf{Our contributions.} In this work, we study the space between these two
extremes. Purely additive NAMs correspond to first-order interaction
structure, where provably minimal explanations are tractable. Fully general
neural networks, by contrast, can encode arbitrary interactions among
features, and in that setting the problem is $\Sigma_2^P$-hard. We study the
intermediate case in which models contain explicit higher-order interactions,
but those interactions remain structured rather than completely unrestricted.
This perspective lets us isolate how the order and connectivity of feature
interactions shape the complexity of computing provable minimal explanations for neural networks.

Our main contributions are as follows:

\begin{itemize}
\item \textbf{Hardness under limited interactions.}
We prove that finding cardinally-minimal provably sufficient explanations for
Any-Order NAMs is $\Sigma_2^P$-hard, even for neural networks over bounded
\emph{constant}-order interactions. Thus, unlike standard NAMs, even limited
interactions make exact provable explanation search intractable.
\item \textbf{Two tractable settings.}
Interestingly, despite our negative complexity result, we recover tractability in two structural settings: (i) a stronger interaction-aware
notion of sufficiency that decouples components; and (ii) sparse interaction
hypergraphs with bounded overlap.
\item \textbf{Certified algorithms for both settings.}
We design neural network verification- and MIP-based algorithms for both settings: the verifier
computes provable component bounds, and the MIP solver searches globally for an optimal minimum over
feature subsets.
\item \textbf{Minimality guarantees.}
In the sparse setting, our method returns minimum-cardinality provably
sufficient explanations for the original model. In the strongly-sufficient explanation
setting, it returns minimum-cardinality explanations under the stronger notion,
which still imply standard sufficiency while retaining significantly smaller explanations in practice.
\item \textbf{Empirical gains.}
We perform extensive experiments across $9$ benchmarks, and show that our methods produce significantly smaller explanations than previously
suggested algorithms, often with better runtime, with models that are significantly more
accurate than standard NAMs. They also provide guarantees unavailable to
sampling-based interpretation methods.

\end{itemize}

Overall, our results significantly expand the class of neural architectures
for which computing provably minimal and sufficient explanations are possible, while
clarifying the precise role of feature interactions in determining their
computational complexity.

Due to space constraints, we present the central ideas behind selected propositions in the main text and defer the full proofs to the appendix.

\section{Preliminaries}

\textbf{Notation.}
Let $[n]:=\{1,\dots,n\}$ denote the feature indices, and let
$\mathcal{D} \subseteq \mathcal{X}_1 \times \cdots \times \mathcal{X}_n$
be the input domain. For ease of presentation, we assume binary
classification, where a score function $F:\mathcal{D}\to\mathbb{R}$ induces
the classifier $h_F(x):=\mathbb{I}[F(x)\geq 0]$; a description of the
regression and multiclass settings is given in Appendix~\ref{app:additional-settings}.
For $x\in\mathcal{D}$ and $S\subseteq[n]$, let $x_S$ denote the restriction of
$x$ to $S$, let $\bar S:=[n]\setminus S$, and define
$\mathcal{C}(x,S):=\{z\in\mathcal{D}:z_S=x_S\}$ as the set of admissible
completions consistent with $x$ on $S$. For any $z\in\mathcal{D}$, we write
$(x_S,z_{\bar S})$ for the hybrid input that agrees with $x$ on $S$ and with
$z$ on $\bar S$.



\subsection{From First-Order to Any-Order NAMs}

Standard (first-order) NAMs decompose the score into a sum of
one-dimensional component functions. This first-order structure makes the
contribution of each feature explicit and completely removes feature
interactions. It is precisely this restriction that underlies the tractability
guarantees known for standard NAM explanations~\citep{AgMeFrZhLeCaHi21, BaMoKa25}.
However, many realistic prediction tasks depend on interactions between
multiple features. Any-Order NAMs extend this additive decomposition by
allowing each component to depend on an arbitrary feature subset rather than on
just one coordinate.

\begin{definition}[Any-Order NAM]
An \emph{Any-Order NAM} has the form
\[
    F(x) := b + \sum_{T \in \mathcal{H}} f_T(x_T),
\]
where $\mathcal{H} \subseteq 2^{[n]} \setminus \{\emptyset\}$ is a family of
interaction scopes and each $f_T$ depends only on the coordinates in $T$
\citep{KiChKi25}. The order of the model is
$\max_{T \in \mathcal{H}}|T|$.

\end{definition}

First-order NAMs are the special case $\max_{T \in \mathcal{H}}|T| = 1$, so
they exclude feature interactions. Any-Order NAMs preserve additivity while
allowing higher-order interactions, making them more expressive and closer to
general neural networks (which allow arbitrary interactions). We ask how much of the tractability of first-order
NAM explanations survives in this richer setting.

\subsection{Neural Network Verification}

Neural network verification asks whether a network satisfies a property
over all inputs in a given set. Formally, given a neural network
$F : \mathcal{D} \to \mathbb{R}$, an input set $\mathcal{X} \subseteq
\mathcal{D}$, and a property $P$, the verification problem is to determine
whether $\forall z \in \mathcal{X},\, P(F,z)$ holds. In classification, the property of interest is often
label preservation. For fixed input $x$ and completion set
$\mathcal{C}(x,S)$, this asks whether
\[
    h_F(z)=h_F(x) \qquad \text{for all } z \in \mathcal{C}(x,S).
\]

This problem is NP-hard in general~\citep{KaBaDiJuKo17, SaLa21}. However, recent years have seen rapid advances in neural network verification, substantially improving the scalability and practical effectiveness of verifiers~\citep{KaLaBaBrDuFlJoKoMaNgWu25, ChChZhZh25, WuIsZeTaDaKoReAmJuBa24}. In particular, verifiers in the CROWN family, such as  the state-of-the-art verifier $\alpha,\beta$-CROWN
\citep{ZhWeChHsDa18, WaZhXuLiJaHsKo21, KaLaBaBrDuFlJoKoMaNgWu25, KoBrKoDvZh23, ZhBrHaZh24}, propagate provable bounds on the value of
each neuron; these bounds can be extracted, and we will use them later.


\section{Provable Explanations for Neural Networks and NAMs}
\label{sec:provable-explanations}

In safety-critical settings, one often seeks explanations with formal
guarantees rather than heuristic scores. For a fixed input, the key question is
whether a subset of observed features preserves the prediction under all
admissible completions of the remaining features, often restricted to a local
neighborhood. For $p \in [1,\infty]$ and $\epsilon > 0$, let
$\mathcal{B}_{p,\epsilon}(x) := \{ z \in \mathcal{D} : \lVert z - x \rVert_p
\leq \epsilon \}$ and
$\mathcal{C}_{p,\epsilon}(x,S):=\mathcal{C}(x,S) \cap \mathcal{B}_{p,\epsilon}(x)$.

\begin{definition}[Sufficient Explanation]
\label{def:local-provably-sufficient-explanation}
For a given neural network $F$, a fixed input $x \in \mathcal{D}$, a norm
parameter $p \in [1,\infty]$, and a perturbation radius $\epsilon > 0$, a set
$S \subseteq [n]$ is a \emph{sufficient explanation} for the query
$\langle F, x, p, \epsilon, S \rangle$ if
\[
    \forall z \in \mathcal{C}_{p,\epsilon}(x,S), \qquad
    h_F(z)=h_F(x).
\]

\end{definition}

Sufficiency alone does not guarantee that an explanation is concise or
informative. The full feature set $[n]$ is always a trivially provably
sufficient explanation.
This motivates the minimality notions below, which seek the
smallest feature subset that still certifies the prediction.

\begin{definition}[Minimal Sufficient Explanations]
\label{def:cardinally-minimal-pse}
\label{def:subset-minimal-pse}
For a given neural network $F$, a fixed input $x \in \mathcal{D}$, a norm
parameter $p \in [1,\infty]$, and a perturbation radius $\epsilon > 0$, a
provably sufficient explanation $S^* \subseteq [n]$ for the query
$\langle F, x, p, \epsilon, S^* \rangle$ may satisfy either of the following
minimality notions:
\begin{enumerate}[label=(\roman*), nosep, leftmargin=*]
\item it is \emph{cardinally-minimal} if $|S^*|\leq |S|$ for every provably
      sufficient explanation $S \subseteq [n]$ for the query
      $\langle F, x, p, \epsilon, S \rangle$;
\item it is \emph{subset-minimal} if no proper subset $S \subsetneq S^*$ is
      itself a provably sufficient explanation for the query
      $\langle F, x, p, \epsilon, S \rangle$.
\end{enumerate}

\end{definition}
These notions capture different goals: cardinal minimality optimizes size
globally, whereas subset minimality only removes redundant revealed features.
Most neural-network explanation methods target the weaker subset-minimal
notion~\citep{BaElLaSaKrAlKa26}, typically via greedy remove-and-verify:
start from the full feature set and delete a feature whenever a verifier still
certifies sufficiency. The result is subset-minimal, but can depend strongly
on the deletion order; sensitivity-based variants differ only in how they
choose that order. We use this baseline later and give pseudocode in
Appendix~\ref{app:background} as Algorithm~\ref{alg:greedy-baseline}.

For general neural networks, computing cardinally-minimal provably sufficient
explanations is already $\Sigma_2^P$-hard, so exact provable minimal
explanations are intractable in full generality; see
Appendix~\ref{app:background} for brief background on this complexity class. NAMs offer an exception. In first-order NAMs, the absence of feature
interactions enables linear-query algorithms for cardinally-minimal provable
explanations~\citep{BaElLaSaKrAlKa26}. This raises the key question of our
work: what happens when interactions are reintroduced? Any-Order NAMs preserve
additivity at the component level while allowing higher-order interactions,
improving expressivity and often accuracy~\citep{KiChKi25}.
Section~\ref{sec:any-order-hardness} shows that interactions restore
hardness, while Sections~\ref{sec:strong-sufficiency}
and~\ref{sec:sparse-any-order} recover tractability under additional
structure.

\section{Hardness of Computing Minimal Explanations for Constant-Order NAMs}
\label{sec:any-order-hardness}

Following previous complexity-theoretic works on sufficient explanations~\citep{BaMoPeSu20, AdViWa25, BaAmKa24, ArBaBaPeSu21, ArBaBeMo23}, we study the decision variant of minimum-size sufficient explanation search: given a budget $B$, decide whether there exists a sufficient explanation of size at most $B$. Formally:


\begin{center}
{\setlength{\fboxsep}{8pt}%
\fbox{%
\begin{minipage}{0.93\linewidth}
\hypertarget{prob:any-order-size-bounded}{}
\textbf{Minimum Sufficient Explanation for $k$-Order NAMs (\sizeboundedproblemname).}
\par\smallskip
Given an Any-Order NAM $F$ of order $k=\max_{T \in \mathcal{H}}|T|$, an input $x \in \mathcal{D}$, a norm
parameter $p \in [1,\infty]$, a perturbation radius $\epsilon > 0$, and an
integer budget $B\in\mathbb{N}$, decide whether there exists a sufficient explanation $S \subseteq [n]$ with respect to $\langle F, x, p, \epsilon, S \rangle$ such
that $|S| \leq B$. 
\end{minipage}}%
}
\end{center}

We show that this problem is already hard even for constant $k$:


\begin{theorem}
\label{thm:any-order-hardness}
There exists a fixed constant $k=\mathcal{O}(1)$ for which \sizeboundedproblemname is
$\Sigma_2^P$-hard.

\end{theorem}

In other words, Theorem~\ref{thm:any-order-hardness} shows that allowing even a constant degree of feature interactions in a NAM raises the complexity of computing cardinally-minimal sufficient explanations to that of fully general neural networks~\citep{BaMoPeSu20}. In particular, the $\Sigma^p_2$-hardness result implies, under standard complexity assumptions, that this task cannot be solved using even a polynomial number of (NP-Complete) neural-network verification queries, and therefore requires \emph{exponentially} many such queries in the worst case. This stands in sharp contrast to first-order neural networks, namely ``classic'' NAMs, where cardinally-minimal sufficient explanations can be computed using a polynomial, and in fact \emph{linear}, number of verification queries~\citep{BaElLaSaKrAlKa26}. Thus, this theorem shows that feature interactions play a central role in the computational hardness of obtaining minimal sufficient explanations for neural networks, and this hardness already appears at very low, constant interaction order.

Interestingly, we show in Sections~\ref{sec:strong-sufficiency}
and~\ref{sec:sparse-any-order} that despite this significantly negative worst-case complexity result,  tractability can nevertheless still be
recovered in practice under one of two assumptions: (i) a stronger interaction-aware notion
of sufficiency or (ii) a sparsity condition on the interaction graph.

\section{A Tractable Relaxation via Strongly Sufficient Explanations}
\label{sec:strong-sufficiency}

Section~\ref{sec:any-order-hardness} shows that, in general, computing
cardinally-minimal sufficient explanations for NAMs of a fixed order is
$\Sigma_2^P$-hard. Thus, in the worst case, exact optimization requires an
exponential number of neural-network verification queries. To overcome this
barrier, we introduce a stronger interaction-aware notion of explanation sufficiency,
which we call \emph{strongly sufficient} explanations. This notion is conservative: a
strongly sufficient explanation is always a sufficient explanation in the
ordinary sense, but it need not be cardinally-minimal among all ordinary
sufficient explanations.

The advantage is computational. Under strong sufficiency, the effect of each
interaction component can be bounded independently, allowing us to compute
provably \emph{cardinally-minimal strongly sufficient explanations} using only a worst-case \emph{linear} number
of verification-query rounds. Although these explanations optimize a stronger
notion than ordinary sufficiency, our experiments show that they are often much
smaller than the explanations produced by standard greedy methods for
sufficient explanations, suggesting that they are also practically closer to
the true ordinary optimum.


\textbf{Strongly sufficient explanations.} Under sufficiency, all omitted features must be completed by a single
globally consistent input. Strong sufficiency removes this coupling and lets
each component choose its own local completion. For a component $T \in \mathcal{H}$ and a set of features $U \subseteq T$,
let
\[
\mathcal{C}^{T}_{p,\epsilon}(x,U) := \{z_T \in \mathcal{D}_T : z_{T \setminus U} = x_{T \setminus U},\ \lVert z_T - x_T \rVert_p \leq \epsilon\}.
\]
For a candidate explanation $S \subseteq [n]$, write $U_T(S) := T \setminus S$.

\begin{definition}[Strongly Sufficient Explanations]
\label{def:strong-sufficiency}
We say that $S$ is a \emph{strongly sufficient explanation} for the query
$\langle F,x,p,\epsilon,S \rangle$ if, for every choice of local completions
$z_T \in \mathcal{C}^{T}_{p,\epsilon}(x,U_T(S))$ for all $T \in \mathcal{H}$,
\[
    \mathbb{I}\!\left[b + \sum_{T \in \mathcal{H}} f_T(z_T) \geq 0\right] = h_F(x).
\]

\end{definition}

\begin{proposition}
\label{prop:strong-implies-sufficient}
For a given Any-Order NAM $F$, a fixed input $x \in \mathcal{D}$, a norm
parameter $p \in [1,\infty]$, a perturbation radius $\epsilon > 0$, and a
candidate explanation $S \subseteq [n]$, if $S$ is strongly sufficient for the
query $\langle F,x,p,\epsilon,S \rangle$, then $S$ is also a sufficient
explanation for the same query.

\end{proposition}

\paragraph{Example.} Let $x_1,x_2,x_3 \in \{0,1\}$, let $F$ have two
components $f_{12}(x_1,x_2)$ and $f_{23}(x_2,x_3)$ (so $b=0$), and consider
$x=(1,0,1)$ with candidate explanation $S=\{1,3\}$ (freeing only $x_2$).
Suppose $f_{12}(1,0){=}{-2}$, $f_{12}(1,1){=}3$, $f_{23}(0,1){=}3$,
$f_{23}(1,1){=}{-2}$. Under ordinary sufficiency, $x_2$ takes one shared
value: both completions ($x_2{=}0$: $-2+3=1$; $x_2{=}1$: $3-2=1$) are
positive, so $S$ is sufficient. Under strong sufficiency, each component
gets its own private copy of $x_2$: taking $x_2{=}0$ in $f_{12}$ and
$x_2{=}1$ in $f_{23}$ gives $-2-2=-4$, so $S$ is \emph{not} strongly
sufficient. An ordinary completion is exactly the special case where the
private copies agree, which is why Proposition~\ref{prop:strong-implies-sufficient} holds.

Thus, strong sufficiency is a conservative relaxation: it may exclude some
ordinary sufficient explanations, but every strongly sufficient explanation is
still certifiably sufficient. The central benefit is that,
once $S$ is fixed, each component can now be analyzed independently through the
free set $U_T(S)$.

\subsection{Finding Minimal Deviations in Each Interaction Component}

To exploit this decoupling, we track how much each component can reduce the
signed score when a subset of its features is left free. Let $\sigma(x) =
2h_F(x)-1 \in \{-1,1\}$. For a component $T \in \mathcal{H}$ and a set $U \subseteq T$, define the
corresponding deviation
$\Delta_T(x,U) := \sigma(x) f_T(x_T) - \min_{z_T \in \mathcal{C}^{T}_{p,\epsilon}(x,U)} \sigma(x) f_T(z_T).$
Thus $\Delta_T(x,U)$ is the worst signed margin loss contributed by component
$T$ when the features in $U$ are allowed to vary. In particular,
$\Delta_T(x,\emptyset)=0$. Because strong sufficiency lets the components choose their completions
independently, the total worst-case margin loss is the sum of these
deviations.

\begin{proposition}
\label{prop:strong-sufficiency-sum}
For a given Any-Order NAM $F$, a fixed input $x \in \mathcal{D}$, a norm
parameter $p \in [1,\infty]$, a perturbation radius $\epsilon > 0$, and a
candidate explanation $S \subseteq [n]$, the set $S$ is strongly sufficient
for the query $\langle F,x,p,\epsilon,S \rangle$ if and only if
\begin{equation}
\sum_{T \in \mathcal{H}} \Delta_T\bigl(x,U_T(S)\bigr) \leq |F(x)|.
\label{eq:strong-sufficiency-sum}
\end{equation}

\end{proposition}

Proposition~\ref{prop:strong-sufficiency-sum} is the key reduction: once the values $\Delta_T(x,U)$ are known, finding a strongly sufficient
explanation becomes a finite combinatorial optimization problem.

\subsection{Finding Cardinally-Minimal Strongly Sufficient Explanations Using MILP}

Assume first that the exact deviation values $\Delta_T(x,U)$ are available for
every component $T \in \mathcal{H}$ for every set $U \subseteq T$. We can
then search for a smallest strongly sufficient explanation with one MILP. We use a binary variable $b_i$ for each feature $i \in [n]$, where $b_i=1$
means that feature $i$ is included in the explanation. We also use a binary
variable $c_{T,U}$ for each pair $(T,U)$, where $c_{T,U}=1$ means that $U$ is
the free set chosen inside component $T$. The exact MILP is
\begin{align}
\text{Objective:}\quad & \min \sum_{i=1}^{n} b_i \label{eq:strong-milp} \\
\intertext{subject to}
& \bullet\ \sum_{U \subseteq T} c_{T,U} = 1
&& \forall T \in \mathcal{H}, \notag \\
& \bullet\ c_{T,U} \leq 1-b_i
&& \forall T \in \mathcal{H},\ \forall U \subseteq T,\ \forall i \in U, \notag \\
& \bullet\ c_{T,U} \leq b_i
&& \forall T \in \mathcal{H},\ \forall U \subseteq T,\ \forall i \in T \setminus U, \notag \\
& \bullet\ \sum_{T \in \mathcal{H}} \sum_{U \subseteq T}
\Delta_T(x,U)c_{T,U} \leq |F(x)|
&& \notag \\
& \bullet\ b_i \in \{0,1\}
&& \forall i \in [n], \notag \\
& \bullet\ c_{T,U} \in \{0,1\}
&& \forall T \in \mathcal{H},\ \forall U \subseteq T. \notag
\end{align}

The first three constraint families ensure that $c_{T,U}$ encodes the free set
induced by the chosen explanation, and the last constraint is exactly the
strong-sufficiency condition from
Proposition~\ref{prop:strong-sufficiency-sum}.

\begin{proposition}
\label{prop:strong-exact-milp}
If \eqref{eq:strong-milp} uses exact deviation values $\Delta_T(x,U)$, every optimal solution encodes a cardinally-minimal strongly sufficient explanation, and vice versa.

\end{proposition}


\textbf{The full algorithm.} Computing every exact deviation $\Delta_T(x,U)$ is computationally infeasible. We therefore use DNN verification to
compute bounds
$\underline{\Delta}_T(x,U) \leq \Delta_T(x,U) \leq \overline{\Delta}_T(x,U)$,
and obtain \emph{optimistic} and \emph{pessimistic} MILP programs by replacing
$\Delta_T(x,U)$ in \eqref{eq:strong-milp} with the lower and upper bounds,
respectively. Appendix~\ref{app:strong-verifier-bounds} gives the details of
the verifier encoding and this MILP construction. If the optimistic and pessimistic MILPs return different objective values, then
some local deviation interval is still too wide. We then use bisection to
narrow the intervals
$[\underline{\Delta}_T(x,U),\overline{\Delta}_T(x,U)]$ for the unresolved
pairs $(T,U)$; Appendix~\ref{app:strong-verifier-bounds} gives the details.

All local pairs $(T,U)$ can be processed independently, so initialization and
refinement are fully parallelizable. Algorithm~\ref{alg:strong-dual-milp}
summarizes the continuous-domain procedure. Appendix~\ref{app:strong-mixed-algorithm}
gives the mixed discrete/continuous variant, where finite completion sets
are handled exactly by enumeration.

\begin{algorithm}[H]
\caption{Strong-sufficiency algorithm for continuous domains}
\label{alg:strong-dual-milp}
\small
\begin{algorithmic}[1]
\Require Any-Order NAM $F$, input $x$, norm parameter $p \in [1,\infty]$, perturbation radius $\epsilon > 0$
\ForAll{$T \in \mathcal{H}$ and $U \subseteq T$ \textbf{in parallel}}
    \State Compute sound bounds $\underline{\Delta}_T(x,U) \leq \Delta_T(x,U) \leq \overline{\Delta}_T(x,U)$
\EndFor
\While{true}
    \State Solve the pessimistic MILP obtained from \eqref{eq:strong-milp} using $\overline{\Delta}_T(x,U)$
    \State Let $S^{\uparrow} \gets \{i \in [n] : b_i^{\uparrow} = 1\}$ and let $v^{\uparrow}$ be its objective value
    \State Solve the optimistic MILP obtained from \eqref{eq:strong-milp} using $\underline{\Delta}_T(x,U)$
    \State Let $v^{\downarrow}$ be its objective value
    \If{$v^{\uparrow} = v^{\downarrow}$}
        \State \Return $S^{\uparrow}$
    \EndIf
    \ForAll{$(T,U)$ with $\underline{\Delta}_T(x,U) < \overline{\Delta}_T(x,U)$ \textbf{in parallel}}
        \State Refine $\underline{\Delta}_T(x,U)$ and $\overline{\Delta}_T(x,U)$ by one bisection step
    \EndFor
\EndWhile
\end{algorithmic}
\end{algorithm}

\begin{theorem}
\label{thm:strong-convergence}
Assume that $\forall T \in \mathcal{H}, U \subseteq T$, the local optimization problem for $\Delta_T(x,U)$ is finitely encoded and verifier calls return rational bounds. Then Algorithm~\ref{alg:strong-dual-milp} terminates after linearly many outer refinement rounds in the encoding size of the Any-Order NAM, and returns a cardinally-minimal strongly sufficient explanation for $\langle F,x,p,\epsilon,S \rangle$. By Proposition~\ref{prop:strong-implies-sufficient}, this set is also sufficient.

\end{theorem}

\textbf{Complexity.}
Each iteration of Algorithm~\ref{alg:strong-dual-milp} makes three calls to an NP oracle: one neural-network verification query and two MILP queries. By Theorem~\ref{thm:strong-convergence}, the number of iterations is linear in the size of the model. Thus, the algorithm makes a \emph{linear} number of NP-oracle calls overall. This stands in sharp contrast to the ordinary sufficiency setting, which is $\Sigma_2^P$-hard (Theorem~\ref{thm:any-order-hardness}) and therefore, under standard complexity assumptions, cannot be solved with polynomially many oracle calls.

\section{A Tractable Relaxation for \texorpdfstring{$k$}{k}-Sparse Any-Order NAMs}
\label{sec:sparse-any-order}

Section~\ref{sec:any-order-hardness} shows that constant-order interactions
are enough to make cardinally-minimal sufficient explanations
$\Sigma_2^P$-hard. Section~\ref{sec:strong-sufficiency} recovered
tractability by strengthening the notion of sufficiency, yielding explanations
that are certified sufficient but minimal only under the stronger notion. Here,
we show that when the interaction structure is sparse, no such relaxation is
needed. By merging overlapping scopes into small disjoint components, ordinary
sufficiency decomposes across components, allowing us to apply the same
linear-query algorithm. Thus, for sparse Any-Order NAMs, we recover a
complexity guarantee matching first-order NAMs: cardinally-minimal sufficient
explanations for the original model can be computed using only a linear number
of verification-query rounds.


\subsection{Merging Interaction Components}

Consider an Any-Order NAM $F(x)=b+\sum_{T \in \mathcal{H}} f_T(x_T)$.
Define the feature-interaction graph $\mathcal{G}_{\mathcal{H}}=([n],E)$ by
placing an edge between two features $i,j \in [n]$ whenever there exists a
scope $T \in \mathcal{H}$ with $i,j \in T$. Let $C_1,\dots,C_m$ be the
connected components of $\mathcal{G}_{\mathcal{H}}$, and call them the
\emph{merged components} of $F$. For each merged component $C_\ell$, define the merged score
$G_\ell(x_{C_\ell}) := \sum_{T \in \mathcal{H}: T \subseteq C_\ell} f_T(x_T)$.
Then the model can be regrouped as $F(x)=b+\sum_{\ell=1}^{m} G_\ell(x_{C_\ell})$.
This preprocessing only changes how the terms are grouped; it does not change
the model.

\begin{lemma}
\label{lem:sparse-merged-components}
Let $C_1,\dots,C_m$ be the merged components of $F$. Then the sets
$C_1,\dots,C_m$ are pairwise disjoint, so each input feature belongs to at most
one merged component. Moreover, every interaction scope $T \in \mathcal{H}$ is
contained in a unique merged component $C_\ell$.

\end{lemma}

\subsection{Cardinally-Minimal Provably Sufficient Explanations in \texorpdfstring{$k$}{k}-Sparse Any-Order NAMs}

\begin{definition}[\texorpdfstring{$s$}{s}-Sparse Any-Order NAM]
\label{def:k-sparse-any-order}
Let $C_1,\dots,C_m$ be the merged components of an Any-Order NAM $F$. We say
that $F$ is \emph{$s$-sparse} if $\max_{\ell \in [m]} |C_\ell| \leq s$. (We
use $s$ rather than $k$ for this bound to keep it distinct from the
interaction order $k$ of an Any-Order NAM.)

\end{definition}

Thus $s$ bounds the largest merged component. When $s$ is small, each score
$G_\ell$ depends on only a few features. We therefore treat each $C_\ell$ as
one component and run Algorithm~1 on the regrouped model
$\widetilde{F}(x)=b+\sum_{\ell=1}^{m} G_\ell(x_{C_\ell})$. No new optimization
machinery is needed: Algorithm~1 simply computes deviations for merged
components rather than original scopes. The mixed discrete/continuous variant
is obtained analogously in Appendix~\ref{app:strong-mixed-algorithm}. By Lemma~\ref{lem:sparse-merged-components}, different merged components do
not interact. Hence $\widetilde{F}$ is exactly the disjoint-component model
handled in Section~\ref{sec:strong-sufficiency}, and strong sufficiency for
$\widetilde{F}$ coincides with ordinary sufficiency for the original model.

\begin{theorem}
\label{thm:sparse-correctness}
Let $F$ be an $s$-sparse Any-Order NAM with merged components
$C_1,\dots,C_m$ and merged scores $G_1,\dots,G_m$. Fix a query
$\langle F,x,p,\epsilon \rangle$. Assume that every local optimization problem
for the merged components is finitely encoded and that each verifier call
returns sound rational bounds. Then applying Algorithm~1 to the regrouped model
$\widetilde{F}$ terminates after a linear number of outer refinement rounds and
returns a cardinally-minimal sufficient explanation for the original model $F$.

\end{theorem}

\textbf{Complexity.} Preprocessing requires only graph connectivity
(e.g., union-find). Theorem~\ref{thm:strong-convergence} then applies to the
regrouped model, with every NP oracle call to a neural network verifier or MILP verifier
restricted to a merged component of size at most $s$. Hence, for constant
$s$, the number of such calls is linear in the model size.

\section{Empirical Evaluation}
\label{sec:empirical-evaluation}

\begingroup
\setlength{\intextsep}{4pt plus 1pt minus 1pt}
\setlength{\textfloatsep}{4pt plus 1pt minus 1pt}
\setlength{\floatsep}{4pt plus 1pt minus 1pt}
\setlength{\abovecaptionskip}{4pt}
\setlength{\belowcaptionskip}{0pt}

\textbf{Experimental setup.} We evaluate the strong-sufficiency algorithm
(Section~\ref{sec:strong-sufficiency}) and the preprocessing-based sparse
reduction (Section~\ref{sec:sparse-any-order}) against two verifier-backed
greedy baselines: canonical-order greedy and a sensitivity-guided variant. All
methods use $\alpha$-$\beta$-CROWN, the state-of-the-art in neural network verification, for the verification queries \citep{ZhWeChHsDa18};
our methods additionally solve optimistic and pessimistic MILPs with Gurobi
\citep{Gu26}, the state-of-the-art MILP solver. Both baselines use greedy remove-and-verify from the full
feature set and differ only in feature order, so they are guaranteed only to
return subset-minimal explanations rather than globally cardinally-minimal
ones. Since all methods share the same
verifier backend, any gains reflect better use of model structure rather than a
weaker certification standard. Exact sample counts, feature counts,
train/test splits, and ablations
appear in Appendix~\ref{app:additional-experimental-details} and
Appendix~\ref{app:ablation-studies}.

\subsection{Provably cardinally-minimal Strongly Sufficient Explanations for Any-Order NAMs}

Table~\ref{tab:strong-sufficiency-results} reports the main strong-sufficiency
results. For each dataset in this experiment, we train a second-order
Any-Order NAM with components for all singleton features and all pairs of input
features. It compares our strongly-sufficient algorithm against these two greedy
subset-minimal baselines in terms of the mean explanation size they produce and
the mean runtime required to compute them. Overall, our method achieves the best mean runtime on
most benchmarks. Mean explanation size is more mixed: since strong
sufficiency is a strictly more conservative notion than ordinary sufficiency
(Proposition~\ref{prop:strong-implies-sufficient}), it carries no guarantee of
beating a subset-minimal baseline on size, and empirically our method attains
the smallest mean size on a minority of benchmarks, remaining close to the
best baseline on the rest. The sparse setting (Section~\ref{sec:sparse-any-order}),
in contrast, does carry such a guarantee and wins on size across the board,
as Table~\ref{tab:sparse-setting-results} below confirms.

\vspace{0.4em}
\begin{table}[H]
\centering
\scriptsize
\setlength{\tabcolsep}{2pt}
\caption{Strong-sufficiency results on nine order-2 tabular benchmarks at $\epsilon=0.1$. Columns show test accuracy, mean explanation size $\pm$ standard deviation (\% of features), and mean runtime $\pm$ standard deviation.}
\label{tab:strong-sufficiency-results}
\resizebox{\linewidth}{!}{%
\begin{tabular}{lccc}
\toprule
Dataset & Acc. & \shortstack{Size mean $\pm$ sd (\%)\\(Ours / greedy / sens.)} & \shortstack{Time mean $\pm$ sd (s)\\(Ours / greedy / sens.)} \\
\midrule
Adult & 0.8635 & \textbf{25.7}$\pm$13.2 / 45.2$\pm$21.7 / 52.8$\pm$29.1 & \textbf{62.943}$\pm$364.322 / 2199.624$\pm$1060.915 / 2267.351$\pm$1467.501 \\
Bike Sharing & 0.9232 & \textbf{33.5}$\pm$17.8 / 36.3$\pm$19.6 / 44.5$\pm$15.7 & \textbf{61.279}$\pm$314.891 / 1137.386$\pm$1351.496 / 2666.257$\pm$1261.087 \\
Breast Cancer & 0.9825 & 4.0$\pm$16.3 / 3.5$\pm$15.9 / \textbf{2.7}$\pm$13.2 & \textbf{126.792}$\pm$447.934 / 3082.496$\pm$402.879 / 3098.785$\pm$365.728 \\
California Housing & 0.8886 & 19.7$\pm$21.8 / 16.7$\pm$19.7 / \textbf{16.0}$\pm$18.8 & 71.860$\pm$412.863 / 18.732$\pm$6.640 / \textbf{17.525}$\pm$5.692 \\
COMPAS & 0.6742 & 50.6$\pm$23.7 / 50.3$\pm$10.1 / \textbf{39.9}$\pm$12.0 & \textbf{26.411}$\pm$113.220 / 2805.141$\pm$1357.731 / 1806.214$\pm$1535.080 \\
Diabetes & 0.7273 & 10.2$\pm$16.6 / 11.1$\pm$19.4 / \textbf{9.8}$\pm$17.0 & \textbf{38.469}$\pm$42.145 / 169.473$\pm$105.444 / 141.862$\pm$79.488 \\
German Credit & 0.7250 & \textbf{64.9}$\pm$14.2 / 77.7$\pm$3.6 / 73.9$\pm$3.1 & \textbf{136.689}$\pm$257.987 / 3550.136$\pm$419.894 / 3569.952$\pm$275.462 \\
Heart Disease & 0.8525 & 29.1$\pm$14.4 / \textbf{27.1}$\pm$10.3 / 34.2$\pm$7.5 & \textbf{130.520}$\pm$54.842 / 2765.516$\pm$1121.928 / 3369.266$\pm$695.254 \\
HELOC & 0.7245 & \textbf{38.4}$\pm$22.1 / 72.0$\pm$5.4 / 77.5$\pm$10.0 & \textbf{910.066}$\pm$557.189 / 3596.063$\pm$251.951 / 3599.857$\pm$186.663 \\
\bottomrule
\end{tabular}}
\end{table}
\vspace{0.4em}
Figure~\ref{fig:strong-sufficiency-curves} shows the COMPAS results; Appendix~\ref{app:additional-strong-sufficiency-curves} gives separate figures for the remaining benchmarks.

\begin{figure}[H]
\centering
\includegraphics[width=0.97\linewidth]{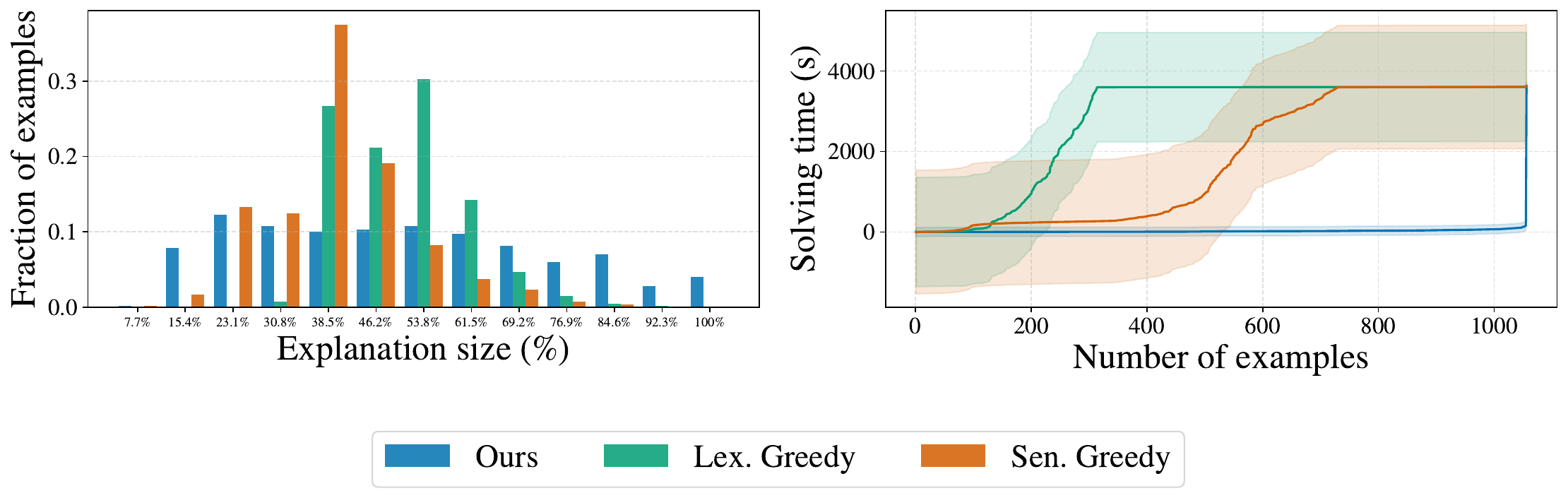}
\caption{COMPAS under strong sufficiency. Combined explanation-size and runtime plot.}
\label{fig:strong-sufficiency-curves}
\end{figure}
\vspace{-1.6em}
\subsection{Provably cardinally-minimal Sufficient Explanations for Sparse Any-Order NAMs}
\label{subsec:sparse-setting-results}

Table~\ref{tab:sparse-setting-results} reports the sparse-setting results. As
in Table~\ref{tab:strong-sufficiency-results}, it compares our sparse
algorithm against the same two greedy baselines, which compute subset-minimal
explanations, in terms of the mean explanation size they find and the mean
runtime required to compute them. In the sparse setting, our preprocessing-based method finds substantially
smaller mean explanation sizes than either greedy baseline across all nine
benchmarks, often by a wide margin, matching the guarantee of
Theorem~\ref{thm:sparse-correctness}; it also achieves the best mean
runtime on most benchmarks, losing only on California Housing, where the
feature set is small enough that greedy's cheaper per-query cost outweighs
our method's advantage in query count.

\vspace{0.4em}
\begin{table}[H]
\centering
\scriptsize
\setlength{\tabcolsep}{2pt}
\caption{Sparse-setting results on nine order-2 tabular benchmarks at $\epsilon=0.1$. Columns show test accuracy, mean explanation size $\pm$ standard deviation (\% of features), and mean runtime $\pm$ standard deviation.}
\label{tab:sparse-setting-results}
\resizebox{\linewidth}{!}{%
\begin{tabular}{lccc}
\toprule
Dataset & Acc. & \shortstack{Size mean $\pm$ sd (\%)\\(Ours / greedy / sens.)} & \shortstack{Time mean $\pm$ sd (s)\\(Ours / greedy / sens.)} \\
\midrule
Adult & 0.8647 & \textbf{19.5}$\pm$16.0 / 25.7$\pm$18.3 / 23.7$\pm$18.8 & \textbf{866.652}$\pm$1397.304 / 1070.071$\pm$994.654 / 1090.491$\pm$951.636 \\
Bike Sharing & 0.9163 & \textbf{23.5}$\pm$13.7 / 29.1$\pm$18.1 / 32.7$\pm$17.0 & \textbf{242.984}$\pm$580.558 / 582.489$\pm$1059.078 / 2215.734$\pm$1224.846 \\
Breast Cancer & 0.9825 & \textbf{1.6}$\pm$7.7 / 2.4$\pm$11.0 / 1.7$\pm$8.0 & \textbf{46.244}$\pm$78.272 / 171.785$\pm$102.700 / 166.080$\pm$97.569 \\
California Housing & 0.8777 & \textbf{15.7}$\pm$18.5 / 16.9$\pm$19.9 / 15.8$\pm$18.6 & 49.813$\pm$201.815 / \textbf{9.818}$\pm$2.837 / 9.802$\pm$2.841 \\
COMPAS & 0.6799 & \textbf{19.7}$\pm$18.5 / 31.5$\pm$20.1 / 23.7$\pm$18.5 & \textbf{40.424}$\pm$33.006 / 1334.135$\pm$1050.061 / 167.761$\pm$260.991 \\
Diabetes & 0.7273 & \textbf{8.8}$\pm$14.2 / 11.6$\pm$18.7 / 9.7$\pm$15.3 & 68.908$\pm$63.193 / 78.631$\pm$58.277 / \textbf{59.267}$\pm$40.367 \\
German Credit & 0.7250 & \textbf{31.3}$\pm$16.2 / 66.3$\pm$8.2 / 54.8$\pm$13.6 & \textbf{215.777}$\pm$258.273 / 3063.644$\pm$822.614 / 1619.714$\pm$1295.282 \\
Heart Disease & 0.8525 & \textbf{18.9}$\pm$10.2 / 24.5$\pm$11.8 / 24.9$\pm$12.7 & \textbf{182.618}$\pm$453.838 / 920.890$\pm$762.224 / 1444.970$\pm$890.476 \\
HELOC & 0.7405 & \textbf{21.1}$\pm$22.0 / 55.2$\pm$14.0 / 62.3$\pm$12.3 & \textbf{710.099}$\pm$866.431 / 2991.469$\pm$1223.455 / 3216.143$\pm$1016.580 \\
\bottomrule
\end{tabular}}
\end{table}
\vspace{0.4em}
Appendix~\ref{app:ablation-studies} additionally compares both settings
against exact cardinally-minimal baselines (brute-force search and a
minimum-hitting-set algorithm) and against VeriX+, a subset-minimal
QuickXplain-based baseline, validates the sparsity assumption of
Section~\ref{sec:sparse-any-order} on interactions automatically detected
from data via SIAN~\citep{EnLi22}, and reports results under the full
(unrestricted) admissible input space rather than an $\epsilon$-ball.

\subsection{Increasing the Interaction Order of the Any-Order NAM}

We use the tabular Bike Sharing benchmark as a case study for demonstrating how our method scales with the interaction order of sparse Any-Order NAMs.
Specifically, we train three sparse models of orders 2, 3, and 4, using
sparsity parameters $s=4,5,6$, respectively, and run the same experiment as in Subsection~\ref{subsec:sparse-setting-results}. Accuracy improves monotonically
with order, from 0.9163 at order 2 to 0.9180 at order 3 and 0.9295 at order
4. Figure~\ref{fig:higher-order-expl-sizes} shows the resulting distribution
of certified explanation sizes at each order; our method solves every
instance at orders 2 and 3, though convergence becomes substantially harder
at order 4 as the largest merged components grow
(Appendix~\ref{app:ablation-studies} reports exact completion rates).

\vspace{0.4em}
\begin{figure}[H]
\centering
\includegraphics[width=0.97\linewidth]{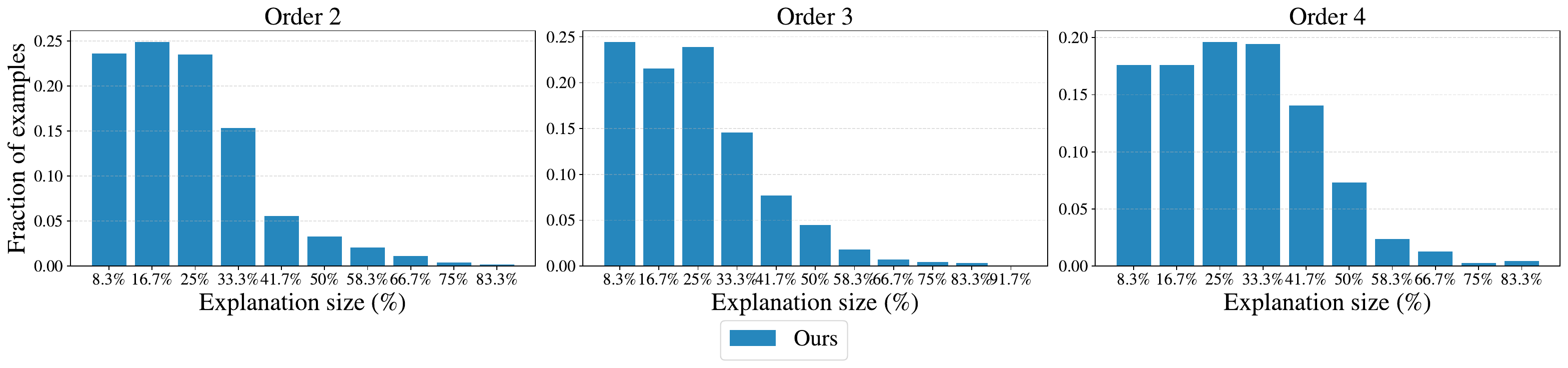}
\caption{Higher-order results for sparse Any-Order NAMs for the Bike Sharing dataset. Explanation-size distributions for orders 2--4, comparing our method with lexicographic greedy and sensitivity-guided greedy search.}
\label{fig:higher-order-expl-sizes}
\end{figure}

\endgroup
\vspace{-1.6em}
\section{Related Work}

\textbf{Formal XAI.}
Our work relates to \emph{formal XAI}~\citep{MaIg22}, which seeks explanations with mathematical guarantees. Prior work on provably minimal sufficient explanations has mainly focused on restricted model classes, including decision trees~\citep{HuIzIgMa21}, tree ensembles~\citep{AuLaMaSz23, AuBeBoKoLaMa22, IzMa21, IgIzStMa22, BoChMaTa21}, linear models~\citep{MaGeCoIgNa20, SuArMe25, BaMoPeSu20}, and monotonic models~\citep{MaGeCoIgNa20, LiHaZhLi20}. Closer to our setting are methods for neural networks~\citep{WuWuBa23, WuLiWuBa26, BaElLaAlKa25, BoElDuBaKa26, IzHuMoPlIgMa24, LaZbMiPaKw21}; however, for general networks, computing cardinally-minimal explanations is computationally hard~\citep{BaMoPeSu20}, so existing approaches typically rely on relaxed or suboptimal variants~\citep{WuWuBa23, BaElLaAlKa25, IzHuMoPlIgMa24}.

For additive models, \citep{BaMoKa25} studied the complexity of several provable explanation notions, including sufficiency, and \citep{BaElLaSaKrAlKa26} proposed a verification-based method for computing cardinally-minimal sufficient explanations for NAMs. However, these approaches exclude feature interactions entirely. In contrast, we are the first to study provable explanations for additive neural models with interactions, where expressivity increases substantially and certification becomes significantly more challenging.

\textbf{Neural Additive Models (NAMs) and feature interactions.} NAMs extend Generalized Additive Models (GAMs)~\citep{HaTi86, NeWe72}, a classic interpretable yet expressive model family~\citep{CaLoGeKoStEl15, ZhChLiSeRu23, LiZhSeRu22, BoVo23, ChWaZhDeHu20}, by replacing each additive component with a neural network, thereby improving expressivity while preserving interpretability. This architecture was first proposed by~\citep{AgMeFrZhLeCaHi21}, with many subsequent extensions in recent years~\citep{RaDuMa22, BoImYeRaFo24, XuBuChBa23, JiZdKiPrWoKiRuShDuMa24, ThKrKnSa24, UpCo24}. In particular, higher-order feature interactions are critical for NAM performance~\citep{EnLi22, RaDuMa22, ChCaGo21}, and have generally received significant attention within the explainability community in recent years~\citep{FuMuKoHuHa23, YaZhSu21, EnLi22, EnLi25}.




\section{Conclusion}

We studied cardinally-minimal provably sufficient explanations for Any-Order
NAMs, showing that feature interactions play a decisive role in determining
when such explanations can be computed efficiently. On the negative side, we prove that even
constant-order interactions recover the $\Sigma_2^P$-hardness of general neural
networks, revealing a sharp computational gap between purely additive NAMs and
models with interactions. On the positive side, we showed that this barrier can
be overcome under additional structure: either by using a stronger
interaction-aware notion of sufficiency, or by exploiting sparsity in the
interaction graph. Together, these results significantly expand the class of neural networks for
which certified explanations can be computed, while clarifying the structural
conditions that make this possible. More broadly, our work highlights feature
interactions as an important potential axis in the theory and practice of
explainability with provable guarantees.

\section*{Acknowledgements and Disclosures}
The work of Refaeli, Elboher, Katz and Bassan was supported by the European Union (ERC, VeriDeL, 101112713).
Views and opinions expressed are however those of the author(s) only and do not necessarily reflect those of the
European Union or the European Research Council Executive Agency.
Neither the European Union nor the granting authority can be held responsible for them.

\bibliography{references}
\bibliographystyle{abbrvnat}

\newpage
\appendix
\onecolumn

\setcounter{theorem}{0}
\renewcommand{\thetheorem}{A.\arabic{theorem}}
\providecommand{\theHtheorem}{}
\renewcommand{\theHtheorem}{appendix.\arabic{theorem}}
\begin{center}\begin{huge} Appendix\end{huge}\end{center}
\noindent{The appendices are organized as follows:}

\newlist{MyIndentedList}{itemize}{4}
\setlist[MyIndentedList,1]{%
	label={},
	noitemsep,
	leftmargin=0pt,
}

\begin{MyIndentedList}
	\item \textbf{Appendix~\ref{limitations}} discusses the main limitations of our
approach, including the dependence on verification or MILP subroutines and the
remaining scalability gap for dense high-order interactions.

	\item \textbf{Appendix~\ref{app:extended-related-work}} discusses extended related work on neural network verification, formal XAI, and additive models with feature interactions.

	\item \textbf{Appendix~\ref{app:background}} contains additional background, including the
greedy-baseline template and polynomial-hierarchy terminology.

	\item \textbf{Appendix~\ref{app:additional-settings}} extends our formulations to other
prediction settings, including general classification and regression.

	\item \textbf{Appendix~\ref{app:hardness-proof}} contains the proof of
Theorem~\ref{thm:any-order-hardness}.

\item \textbf{Appendix~\ref{app:omitted-proofs}} contains the proofs of all other theorems,
propositions, and lemmas in the main text.

\item \textbf{Appendix~\ref{app:additional-experimental-details}} contains additional
experimental details, reproducibility information, compute requirements, and dataset summaries.

\item \textbf{Appendix~\ref{app:ablation-studies}} contains ablation studies, including
\texorpdfstring{$\epsilon$}{epsilon} sensitivity and order-comparison sweeps.

\end{MyIndentedList}



\section{Limitations}
\label{limitations}

Like all other methods for computing provably minimal sufficient explanations,
our approach relies on neural-network verification queries. Although the
scalability of neural-network verification has improved rapidly in recent
years~\citep{KaLaBaBrDuFlJoKoMaNgWu25, ChChZhZh25, WuIsZeTaDaKoReAmJuBa24, ZhWeChHsDa18, WaZhXuLiJaHsKo21, KaLaBaBrDuFlJoKoMaNgWu25, KoBrKoDvZh23, ZhBrHaZh24}, existing verifiers still do not scale to state-of-the-art neural
networks. Thus, our results should be viewed as expanding the tractable frontier
of certified explainability, rather than removing the underlying verification
bottleneck. As neural network verification tools continue
to improve, the scalability of our approach is expected to improve as well.

Our main contribution is therefore not to eliminate verification cost, but to
show that, under suitable interaction structure, the number of required
verification queries can be reduced from worst-case exponential to linear. This
makes provable explanation search practical for richer models with feature
interactions, which achieve substantially higher accuracy than standard NAMs
and yield significantly smaller explanations than greedy verifier-backed
baselines across standard benchmarks.

This reliance on verification and optimization solvers is not an
implementation choice but a consequence of Theorem~\ref{thm:any-order-hardness}:
since computing a provably cardinally-minimum sufficient explanation is
already $\Sigma_2^P$-hard for constant-order Any-Order NAMs, obtaining such
explanations necessarily involves solver calls (assuming $\mathrm{P}\neq\mathrm{NP}$). Two solvers play distinct roles. Provable sufficiency requires
certifying that no admissible completion of the free features can flip the
prediction; over a continuous region this requires a sound and complete
neural-network verifier, whereas for discrete features with enumerable
domains the maximal attainable deviation can instead be obtained directly by
enumeration, with no verifier needed. Separately, even when every feature is
discrete and all deviations are known exactly, a MILP solver (or an
equivalent combinatorial search) is still required to extract the
cardinally-minimal explanation itself; a heuristic such as our greedy
baselines can substitute for either step, but only at the cost of the
minimality guarantee.

Finally, our exact guarantees are specific to Any-Order NAMs rather than
arbitrary (non-additive) neural architectures; for general DNNs, a
minimum-cardinality sufficient explanation is $\Sigma_2^P$-hard to compute
even to find~\citep{BaMoPeSu20}, so an efficient exact extension to that
setting is unlikely. One practical bridge is to first train a DNN, detect
its significant interactions, fit a sparse Any-Order NAM to exactly those
interactions, and then apply our method to the resulting model -- the SIAN
pipeline of Appendix~\ref{app:sian} -- which in our experiments remains
close to the original DNN's accuracy while admitting an efficient certified
explanation. Training DNNs with objectives that explicitly favor low-order
interaction structure, and using automatically discovered interactions as a
heuristic for existing (uncertified) explanation methods for general DNNs,
are both directions we leave to future work.

\section{Extended Related Work}
\label{app:extended-related-work}

\textbf{Neural network verification.} Neural network verification provides provable guarantees about the behavior of a network over a continuous input region, and is NP-complete even for simple ReLU networks~\citep{KaBaDiJuKo17, SaLa21}. Classical SMT- and MILP-based approaches~\citep{KaBaDiJuKo17, KaHuIbJuLaLiShThWuZeDiKoBa19, WuIsZeTaDaKoReAmJuBa24, Eh17, TjXiTe19} encode the network exactly, yielding exact guarantees but scaling only to small-to-medium models, whereas abstract interpretation methods~\citep{GeMiDrTsChVe18, SiGePuVe19, MuMaSiPuVe22} propagate layer-wise over-approximations that are fast but incomplete. Linear bound propagation, notably CROWN~\citep{ZhWeChHsDa18}, and the branch-and-bound frameworks built on it~\citep{WaZhXuLiJaHsKo21, FeMuJoVe22, ZhBrHaZh24, ShJiKoJaHsZh25, ChChZhZh25, KoBrKoDvZh23} now achieve complete verification at much larger scales, as tracked by VNN-COMP~\citep{BrBaJoWu24, KaLaBaBrDuFlJoKoMaNgWu25}. We use such verifiers only as subroutines that bound individual interaction components, combined through a MILP~\citep{Gu26}, so our approach directly benefits from future verifier improvements.

\textbf{Formal XAI.} Our work belongs to formal XAI~\citep{MaIg22, Ma23, Da23}, which studies explanations with provable guarantees, such as sufficient reasons, also known as abductive or prime-implicant explanations~\citep{ShChDa18, IgNaMa19, DaHi20, DaJi22, YuIgStNaMa23, AuBeBoKoLaMa21}, typically computed with tools such as QuickXplain and MaxSAT~\citep{Ju04, IgMoMa19}. Because such explanations are often computationally intractable~\citep{BaMoPeSu20, ArBaBaPeSu21, BaAmKa24, AmBaKa24, BaAmZeKa25, BaHuKa26, AdViWa25, ArBaRoSu22, BlLaTa21, BlKoLaTa22}, a pattern that also holds for SHAP~\citep{VaLyScSu21, ArBaBeMo23, MaDe24, MaBaKaDe25, MaBaKa25}, much of the literature focuses on restricted model classes~\citep{MaIg23}, including decision trees~\citep{HuIzIgMa21, BoKo23, Bo24, Bo25}, tree ensembles~\citep{IzMa21, IgIzStMa22, AuBeBoKoLaMa22, AuLaMaSz23, BoChMaTa21}, linear models~\citep{MaGeCoIgNa20, SuArMe25}, monotonic models~\citep{MaGeCoIgNa21, ElCuHo23, LiHaZhLi20}, graph neural networks~\citep{AzMaPaTe25}, and additive models~\citep{BaMoKa25, BaElLaSaKrAlKa26}. Closer to our setting are methods for general neural networks~\citep{IgNaMa19, LaZbMiPaKw21, BaKa23, BaAmCoReKa23, WuWuBa23, WuLiWuBa26, IzHuMoPlIgMa24}, including several recent works that improve scalability or broaden the form of the explanation~\citep{DeDoUr26, LaKoBlFeWaSh25, SoChGiGrXuCa26, WaMaRoOgJoMa26, HaKaBa26}. Since computing cardinally-minimal explanations for general networks is $\Sigma_2^P$-hard~\citep{BaMoPeSu20}, other approaches alleviate this cost through abstractions~\citep{BaElLaAlKa25, BoElDuBaKa26}, smoothing~\citep{XuAlWo23, JiXuYoGoWo25, AnLoFrSc25}, or self-explaining architectures~\citep{AlJa18, BaElGu25, BaGuZeMaElKy26, YoQuGaJaWo25}. Verification has also been used to equip feature attributions with guarantees, through verified perturbation analysis~\citep{FeDuViCaCaNiSe23} and provable bounds on exact SHAP values~\citep{LuLe17} of neural networks~\citep{BoBaKaLeSu26}.

\textbf{Feature interactions and additive models.} Feature interactions have been widely studied in explainability, through Shapley interaction indices~\citep{GrRo99, SuDhAg20, TsYeRa23, FuMuKoHuHa23}, their connections to GAMs~\citep{BoVo23, EnLi25}, and interaction detection in trained networks~\citep{TsChLi18}. On the modeling side, GAMs~\citep{HaTi86} were extended with interactions in GA$^2$Ms~\citep{LoCaGeHo13, CaLoGeKoStEl15} and in many neural variants~\citep{YaZhSu21, ChCaGo21, RaDuMa22, DuRaMa22, EnLi22, IbAfBeMa23, KiChKi25}, which typically keep the set of interactions sparse or structured to retain interpretability. Our results give this design choice a complexity-theoretic justification: sparse, bounded-overlap interaction structure is one of the settings under which provably cardinally-minimal sufficient explanations become tractable.

\section{Additional Background}
\label{app:background}

\subsection{Greedy baseline template}

For convenience, we record the generic verifier-backed greedy template used by
the baselines in Section~\ref{sec:empirical-evaluation}. The canonical and
sensitivity-based variants differ only in the ordering routine $\pi$.

\begin{algorithm}[H]
\caption{Generic greedy remove-and-verify baseline}
\label{alg:greedy-baseline}
\small
\begin{algorithmic}[1]
\Require query $\langle F, x, p, \epsilon \rangle$, ordering routine $\pi$
\Ensure subset-minimal provably sufficient explanation $S$
\State $S \gets [n]$
\State $(i_1,\dots,i_n) \gets \pi(F, x, p, \epsilon)$
\For{$t = 1, \dots, n$}
    \State $S' \gets S \setminus \{i_t\}$
    \If{$S'$ is verified sufficient for $\langle F, x, p, \epsilon, S' \rangle$}
        \State $S \gets S'$
    \EndIf
\EndFor
\State \Return $S$
\end{algorithmic}
\end{algorithm}

\subsection{Polynomial-hierarchy terminology}

\begin{definition}[\texorpdfstring{$\Sigma_2^P$}{Sigma2P}]
\label{def:sigma2p}
A decision problem $L$ belongs to $\Sigma_2^P$ (also written $\Sigma^P_2$) if
there exist a polynomial $p$ and a polynomial-time computable predicate $R$
such that for every instance $w$,
\[
    w \in L
    \quad\Longleftrightarrow\quad
    \exists u \in \{0,1\}^{p(|w|)}\ \forall v \in \{0,1\}^{p(|w|)}:
    R(w,u,v)=1.
\]
Equivalently, $\Sigma_2^P = \mathrm{NP}^{\mathrm{NP}}$, namely the class of
decision problems solvable by an NP machine with access to an NP oracle.

\end{definition}

\begin{definition}[\texorpdfstring{$\Sigma_2^P$}{Sigma2P}-Hardness]
\label{def:sigma2p-hard}
A decision problem is \emph{$\Sigma_2^P$-hard} if every problem in
$\Sigma_2^P$ can be reduced to it by a polynomial-time many-one reduction.
Intuitively, such a hardness result means that the problem is at least as hard
as the canonical quantified problems at the second level of the polynomial
hierarchy. In particular, an exact polynomial-time algorithm for a
$\Sigma_2^P$-hard problem would imply a collapse of the polynomial hierarchy.

\end{definition}

\section{Extensions to Additional Architectures}
\label{app:additional-settings}

The main text focuses on binary classification, where a single score function
$F$ induces the prediction $h_F(x)=\mathbb{I}[F(x)\geq 0]$. As in the
strategy extends to multi-class classification and regression after modifying
the target margin condition.

\subsection{General Classification}

Suppose the model has class scores $F^{(1)},\dots,F^{(c)}$ and predicts the
label
\[
    t = \arg\max_{j \in [c]} F^{(j)}(x).
\]
To preserve this prediction, it suffices to enforce for every admissible
completion $z \in \mathcal{C}(x,S)$ and every rival class $j \neq t$ that
\[
    F^{(t)}(z) - F^{(j)}(z) \geq 0.
\]
Equivalently, one may work with pairwise winner-vs.-one margins
\[
    M_j(x) = F^{(t)}(x) - F^{(j)}(x), \qquad j \neq t,
\]
or with the winner-vs.-all margin
\[
    M_{\mathrm{all}}(x) = F^{(t)}(x) - \max_{j \neq t} F^{(j)}(x).
\]
For Any-Order NAMs, each pairwise margin $M_j$ is again an Any-Order NAM after
merging the component families of the two logits, so the binary machinery from
Sections~\ref{sec:strong-sufficiency} and~\ref{sec:sparse-any-order} applies
without conceptual changes. In the strong-sufficiency formulation, one simply
replaces the single binary margin constraint by the family of constraints
associated with all rivals $j \neq t$; in the sparse setting, the same
regrouping argument is then applied to each pairwise margin model. Any subset
that satisfies all of these pairwise constraints preserves the original winning
class under every admissible completion.

\subsection{Regression}

For regression, a natural analogue is to fix a tolerance $\delta > 0$ and call
$S$ sufficient if every admissible completion keeps the output within $\delta$
of the original prediction:
\[
    |F(z) - F(x)| \leq \delta \qquad \text{for all } z \in \mathcal{C}(x,S).
\]
This is equivalent to the pair of one-sided constraints
\[
    F(z) - F(x) \leq \delta
    \qquad\text{and}\qquad
    F(x) - F(z) \leq \delta.
\]
Accordingly, in the strongly sufficient setting we compute, for each component
$T \in \mathcal{H}$ and free set $U \subseteq T$, both the largest upward
and downward deviations,
\[
    \Delta_T^{+}(x,U)
    := \max\{ f_T(z_T) - f_T(x_T) : z_T \text{ is feasible and } z_U = x_U \},
\]
and
\[
    \Delta_T^{-}(x,U)
    := \max\{ f_T(x_T) - f_T(z_T) : z_T \text{ is feasible and } z_U = x_U \}.
\]
The MILP then replaces the single classification-margin constraint with
\[
    \sum_{T \in \mathcal{H}} \Delta_T^{+}(x,U_T(S)) \leq \delta
    \qquad\text{and}\qquad
    \sum_{T \in \mathcal{H}} \Delta_T^{-}(x,U_T(S)) \leq \delta.
\]
The sparse-setting method extends analogously after regrouping into merged
components. Thus, once the desired regression guarantee is expressed through an
output tolerance, the same verifier-and-MILP template yields certified
minimum-cardinality explanations.

\subsection{Contrastive Explanations}

A contrastive explanation is a minimal set of features that, if changed,
can flip the prediction; it is the dual notion to a sufficient explanation,
since a set is contrastive exactly when its complement is \emph{not}
sufficient. We ask whether our two tractable settings dualize to yield
cardinally-minimal contrastive explanations as well.

\textbf{Sparse setting (yes).} Let $C_1,\dots,C_m$ be the merged components
of an $s$-sparse Any-Order NAM (Definition~\ref{def:k-sparse-any-order}). By
Proposition~\ref{prop:strong-sufficiency-sum} applied to the regrouped model
$\widetilde{F}$, a candidate $C \subseteq [n]$ is contrastive iff
$\sum_{\ell=1}^{m} \Delta_{C_\ell}\bigl(x, C_\ell \cap C\bigr) > |F(x)|$, the
complement of the sufficiency criterion. Since the merged components are
disjoint (Lemma~\ref{lem:sparse-merged-components}), each component's worst
case is attained on its own coordinates, so the decoupled sum remains exact
and any violation of the margin is witnessed by a single real perturbation
(not merely a family of locally inconsistent ones). Swapping the free and
fixed indicator constraints in the MILP of Section~\ref{sec:sparse-any-order}
and swapping the deviation bounds in Algorithm~1 yields an analog of
Theorem~\ref{thm:sparse-correctness} for cardinally-minimal contrastive
explanations, in the same linear number of outer refinement rounds.

\textbf{Strong sufficiency setting (no).} The two notions relax in opposite
directions. Strong sufficiency is conservative: decoupling the components'
completions only enlarges the set of adversaries considered, so a strongly
sufficient set remains genuinely sufficient
(Proposition~\ref{prop:strong-implies-sufficient}). Contrastiveness has no
such safety margin in the other direction: a decoupled witness to the
contrastive analog of~\eqref{eq:strong-sufficiency-sum} is a family of local
completions, one per component, that need not agree on the value of any
feature they share, and so need not correspond to any single admissible
input. A set that is contrastive under this decoupled notion therefore need
not be contrastive under the ordinary definition, which is why we do not
dualize the strong-sufficiency algorithm of
Section~\ref{sec:strong-sufficiency}; checking a candidate there instead
requires a direct query on the original (non-decoupled) model asking
whether some admissible perturbation of it flips the prediction. A provably
minimum-cardinality contrastive algorithm for the strong-sufficiency
setting is left to future work.

\section{Proof of Theorem~\ref{thm:any-order-hardness}}
\label{app:hardness-proof}

In this appendix we prove Theorem~\ref{thm:any-order-hardness}: there exists a
fixed constant $k$ such that \sizeboundedproblemname, restricted to Boolean
domains and Any-Order NAMs of order at most $k$, is $\Sigma_2^P$-hard. Thus we
work throughout with Boolean domains, $\mathcal{D} = \{0,1\}^n$, and
interaction scopes of size at most $k$. For simplicity, we also fix
$p=\infty$ and $\epsilon=1$; then every Boolean point lies in
$\mathcal{B}_{p,\epsilon}(x)$, so $\mathcal{C}_{p,\epsilon}(x,S)=\mathcal{C}(x,S)$.
A standard route for extending such Boolean hardness constructions to general
continuous domains is described in~\citep{SaLa21}. Since this
restricted family is a special case of Any-Order NAMs, the theorem from the
main text follows immediately.

We begin with the logic-side problem from which the reduction is derived.
A literal over a variable $x_i$ is either $x_i$ or $\neg x_i$. For an
assignment $x \in \{0,1\}^n$, we write $x \models \ell$ when the literal
$\ell$ evaluates to true under $x$. If
$u = \ell_{i_1} \wedge \cdots \wedge \ell_{i_q}$
is a conjunction of literals, then $x \models u$ means that
$x \models \ell_{i_j}$ for every $j \in [q]$. A DNF formula is a formula of
the form
\[
    \psi = u_1 \vee \cdots \vee u_m,
\]
where each $u_j$ is a conjunction of literals. A set of literals $C$ is an
\emph{implicant} of $\psi$ if every total assignment satisfying all literals in
$C$ also satisfies $\psi$.

The reduction uses a restricted constant-width Shortest Implicant Core problem.
Fix a constant $d \in \mathbb{N}$. An instance consists of a DNF formula
\[
    \psi = u_1 \vee \cdots \vee u_m \vee t^{\star}
\]
over variables $x_1,\dots,x_n$ together with an integer $r$, such that:
(i) every term of $\psi$ has width at most $d$, and (ii) every variable
appearing in $\psi$ also appears in the distinguished term $t^{\star}$.
The question is whether there exists an implicant
$C \subseteq t^{\star}$ of $\psi$ such that $|C| \leq r$.
By combining the constant-width SIC hardness route with the DNF-to-DNF
purification theorem in~\citep{BaElLaSaKrAlKa26}, one obtains
that this restricted problem is $\Sigma_2^P$-hard for some fixed constant $d$.
We use that fact as an imported ingredient.

\begin{lemma}
\label{lem:term-component}
Let $u = \ell_{i_1} \wedge \cdots \wedge \ell_{i_q}$ be a conjunction of
literals of width $q$. Then there exists a component function
$g_u : \{0,1\}^{\{i_1,\dots,i_q\}} \to \{0,1\}$ such that for every
$x \in \{0,1\}^n$,
\[
    g_u(x_{\{i_1,\dots,i_q\}}) = 1 \iff x \models u.
\]

\end{lemma}

\paragraph{Proof.}
Define $g_u(y)=1$ if and only if the partial assignment $y$ satisfies every
literal appearing in $u$, and set $g_u(y)=0$ otherwise. By construction,
$g_u$ depends only on the variables occurring in $u$, and it evaluates to $1$
exactly on the satisfying assignments of $u$. \hfill$\square$

\begin{lemma}
\label{lem:dnf-to-any-order-nam}
Let
\[
    \psi = u_1 \vee \cdots \vee u_m
\]
be a DNF in which every term has width at most $k$. Then there exists an
Any-Order NAM of order at most $k$ with score function $G_{\psi}$ such that for
every $x \in \{0,1\}^n$,
\[
    h_{G_{\psi}}(x) = 1 \iff x \models \psi.
\]

\end{lemma}

\paragraph{Proof.}
For each term $u_j$, let $g_{u_j}$ be the component from Lemma~\ref{lem:term-component}, and define
\[
    G_{\psi}(x) = -\frac{1}{2} + \sum_{j=1}^{m} g_{u_j}(x_{T_j}),
\]
where $T_j$ is the variable set appearing in $u_j$. Every component has arity
at most $k$, so this is an Any-Order NAM of order at most $k$. If $x \models
\psi$, then at least one term is satisfied and hence
$\sum_{j=1}^{m} g_{u_j}(x_{T_j}) \geq 1$, implying
$G_{\psi}(x) \geq \frac{1}{2}$ and therefore $h_{G_{\psi}}(x)=1$. If
$x \not\models \psi$, then every term is false, so each component contributes
$0$, yielding $G_{\psi}(x)=-\frac{1}{2}$ and therefore
$h_{G_{\psi}}(x)=0$. \hfill$\square$

\begin{lemma}
\label{lem:implicant-to-explanation}
Let
\[
    \psi = u_1 \vee \cdots \vee u_m \vee t^{\star}
\]
be a DNF over variables $x_1,\dots,x_n$ such that every variable appearing in
$\psi$ also appears in $t^{\star}$. Let $G_{\psi}$ be the score function from
Lemma~\ref{lem:dnf-to-any-order-nam}, and let $x^{\star} \in \{0,1\}^n$ be the assignment satisfying all
literals of $t^{\star}$. For any set of literals $C \subseteq t^{\star}$, define
\[
    S_C = \{ i \in [n] : \text{the literal of $t^{\star}$ over $x_i$ belongs to $C$} \}.
\]
Then
\[
    C \text{ is an implicant of } \psi
    \iff
    S_C \text{ is a sufficient explanation for } x^{\star}
    \text{ with respect to } G_{\psi}.
\]

\end{lemma}

\paragraph{Proof.}
By Lemma~\ref{lem:dnf-to-any-order-nam}, the classifier $h_{G_{\psi}}$ computes exactly the Boolean
function represented by $\psi$.

Suppose first that $C$ is an implicant of $\psi$. Let
$z \in \mathcal{C}(x^{\star},S_C)$ be any completion consistent with $x^{\star}$
on $S_C$. By definition of $S_C$, the point $z$ satisfies all literals in $C$.
Since $C$ is an implicant, this implies $z \models \psi$, and hence
$h_{G_{\psi}}(z)=1$. Because $x^{\star}$ satisfies $t^{\star}$, it also
satisfies $\psi$, so $h_{G_{\psi}}(x^{\star})=1$. Therefore
$h_{G_{\psi}}(z)=h_{G_{\psi}}(x^{\star})$ for every
$z \in \mathcal{C}(x^{\star},S_C)$, and $S_C$ is a sufficient explanation.

Conversely, suppose that $S_C$ is a sufficient explanation for $x^{\star}$, and
let $z \in \{0,1\}^n$ satisfy all literals in $C$. Because every variable of
$\psi$ appears in $t^{\star}$, satisfying the literals in $C$ is equivalent to
agreeing with $x^{\star}$ on the coordinates in $S_C$. Thus
$z \in \mathcal{C}(x^{\star},S_C)$. Since $S_C$ is sufficient,
$h_{G_{\psi}}(z)=h_{G_{\psi}}(x^{\star})=1$. Using again that $h_{G_{\psi}}$
computes $\psi$ exactly, we conclude that $z \models \psi$. Thus every total
assignment satisfying $C$ also satisfies $\psi$, so $C$ is an implicant of
$\psi$. \hfill$\square$

\paragraph{Proof of Theorem~\ref{thm:any-order-hardness}.}
Throughout this proof, we assume that all activations in the network are
ReLUs. Fix a constant $d$ for which the restricted SIC problem above is
$\Sigma_2^P$-hard, and choose any fixed $k \geq d$. We fix the query
parameters to $p=\infty$ and $\epsilon=1$, and reduce from that restricted
SIC problem to the restriction of \sizeboundedproblemname to Boolean-domain
Any-Order NAMs of order at most $k$.

Consider an input instance $(\psi,r)$, where
\[
    \psi = u_1 \vee \cdots \vee u_m \vee t^{\star}
\]
and every term has width at most $d$. Construct the order-$k$ Any-Order NAM
with score function $G_{\psi}$ from Lemma~\ref{lem:dnf-to-any-order-nam}, and let
$x^{\star} \in \{0,1\}^n$ be the assignment satisfying all literals of the
distinguished term $t^{\star}$. We claim that
\[
\begin{aligned}
    (\psi,r) &\text{ is a yes-instance of restricted SIC} \\
    &\iff (G_{\psi},x^{\star},p,\epsilon,r) \\
    &\qquad\text{ is a yes-instance of \sizeboundedproblemname}.
\end{aligned}
\]

For the forward direction, suppose there exists an implicant
$C \subseteq t^{\star}$ with $|C| \leq r$. Let $S_C$ be the corresponding
feature set from Lemma~\ref{lem:implicant-to-explanation}. That lemma implies that $S_C$ is a sufficient
explanation for $x^{\star}$ with respect to $G_{\psi}$, and clearly
$|S_C|=|C| \leq r$. Hence $(G_{\psi},x^{\star},p,\epsilon,r)$ is a yes-instance of
\sizeboundedproblemname.

For the reverse direction, suppose $(G_{\psi},x^{\star},p,\epsilon,r)$ is a yes-instance
of \sizeboundedproblemname. Then there exists a sufficient explanation $S \subseteq [n]$
with $|S| \leq r$. Because every variable of $\psi$ appears in $t^{\star}$, the
set $S$ determines a subset of literals
\[
    C_S = \{ x_i : i \in S,\ x_i^{\star}=1 \}
    \cup
    \{ \neg x_i : i \in S,\ x_i^{\star}=0 \},
\]
which is contained in $t^{\star}$ and satisfies $|C_S|=|S| \leq r$.
Applying Lemma~\ref{lem:implicant-to-explanation} in the reverse direction shows that $C_S$ is an implicant
of $\psi$. Thus $(\psi,r)$ is a yes-instance of the restricted SIC problem.

The reduction is polynomial-time: for each DNF term we create one NAM
component, and each component has arity at most the fixed constant $k$.
Therefore \sizeboundedproblemname is $\Sigma_2^P$-hard even for Boolean-domain Any-Order
NAMs of order at most $k$, and hence also for unrestricted Any-Order NAMs.
\hfill$\square$

\section{Omitted Proofs}
\label{app:omitted-proofs}

For brevity, throughout this appendix we use the signed-score formulation from
Sections~\ref{sec:strong-sufficiency} and~\ref{sec:sparse-any-order}: the
relevant certification condition is that the signed score remains nonnegative
on the completion set under consideration. We also abbreviate
$U_T(S)=T \cap \bar{S}$ for $T \in \mathcal{H}$ and
$W_\ell(S)=C_\ell \cap \bar{S}$ for sparse blocks.

\subsection{Proofs for Section~\ref{sec:strong-sufficiency}}

\paragraph{Proof of Proposition~\ref{prop:strong-implies-sufficient}.}
Assume that $S \subseteq [n]$ is strongly sufficient for the query
$\langle F,x,p,\epsilon,S \rangle$, and let
$z \in \mathcal{C}_{p,\epsilon}(x,S)$ be any admissible completion consistent
with $x_S$. For each component $T \in \mathcal{H}$, define $y^T = z_T$. Since
$z_S = x_S$, we have
\[
    y^T_{S \cap T} = z_{S \cap T} = x_{S \cap T}
    \qquad \text{for every } T \in \mathcal{H}.
\]
Moreover, because $\|z-x\|_p \leq \epsilon$, every subvector also satisfies
$\|y^T-x_T\|_p = \|z_T-x_T\|_p \leq \epsilon$.
Therefore $(y^T)_{T \in \mathcal{H}} \in \mathfrak{C}^{\mathrm{str}}(x,S)$.
By strong sufficiency,
\[
    \mathbb{I}\!\left[b + \sum_{T \in \mathcal{H}} f_T(y^T) \geq 0\right] = h_F(x).
\]
But
\[
    b + \sum_{T \in \mathcal{H}} f_T(y^T)
    = b + \sum_{T \in \mathcal{H}} f_T(z_T)
    = F(z),
\]
so $h_F(z)=h_F(x)$. Since $z \in \mathcal{C}_{p,\epsilon}(x,S)$ was arbitrary,
$S$ is a sufficient explanation for the query
$\langle F,x,p,\epsilon,S \rangle$.
\hfill$\square$

\paragraph{Proof of Proposition~\ref{prop:strong-sufficiency-sum}.}
Fix a candidate set $S \subseteq [n]$. For each component $T \in \mathcal{H}$,
define
\[
    m_T(S) = \min\Bigl\{ \sigma(x)f_T(z_T) : z_T \in \mathcal{D}_T,
    \ z_{T \setminus U_T(S)} = x_{T \setminus U_T(S)},\
    \ \lVert z_T - x_T \rVert_p \leq \epsilon \Bigr\}.
\]
By definition of the component deviation $\Delta_T$ (Section~\ref{sec:strong-sufficiency}),
\[
    \Delta_T\bigl(x,U_T(S)\bigr) = \sigma(x)f_T(x_T) - m_T(S).
\]

Because strong sufficiency allows each component to choose its completion
independently, the set $\mathfrak{C}^{\mathrm{str}}(x,S)$ is a Cartesian
product over the components. Therefore, the minimum attainable signed score over
all strong completions factorizes as
\begin{align*}
    \min_{(y^T)_{T \in \mathcal{H}} \in \mathfrak{C}^{\mathrm{str}}(x,S)}
    \sigma(x)\Bigl(b + \sum_{T \in \mathcal{H}} f_T(y^T)\Bigr)
    &= \sigma(x)b + \sum_{T \in \mathcal{H}} m_T(S).
\end{align*}
Also,
\[
    |F(x)| = \sigma(x)b + \sum_{T \in \mathcal{H}} \sigma(x)f_T(x_T).
\]
Subtracting the two identities yields
\begin{align*}
    |F(x)| - \sum_{T \in \mathcal{H}} \Delta_T\bigl(x,U_T(S)\bigr)
    &= \sigma(x)b + \sum_{T \in \mathcal{H}} \sigma(x)f_T(x_T)
      - \sum_{T \in \mathcal{H}} \bigl(\sigma(x)f_T(x_T) - m_T(S)\bigr) \\
    &= \sigma(x)b + \sum_{T \in \mathcal{H}} m_T(S) \\
    &= \min_{(y^T) \in \mathfrak{C}^{\mathrm{str}}(x,S)}
       \sigma(x)\Bigl(b + \sum_{T \in \mathcal{H}} f_T(y^T)\Bigr).
\end{align*}

Hence $S$ is strongly sufficient exactly when the minimum signed score over
$\mathfrak{C}^{\mathrm{str}}(x,S)$ is nonnegative, which is equivalent to
\[
    |F(x)| - \sum_{T \in \mathcal{H}} \Delta_T\bigl(x,U_T(S)\bigr) \geq 0.
\]
Rearranging gives \eqref{eq:strong-sufficiency-sum}. \hfill$\square$

\subsubsection{Bounding minimal deviations with DNN verification}
\label{app:strong-verifier-bounds}

Computing every exact value $\Delta_T(x,U)$ requires solving a nonlinear
optimization problem over $\mathcal{C}^{T}_{p,\epsilon}(x,U)$. To obtain
certified bounds instead, we encode the local component network $f_T$ together
with the constraints defining $\mathcal{C}^{T}_{p,\epsilon}(x,U)$ as a verifier
query. Concretely, the coordinates in $T \setminus U$ are fixed to their values
in $x$, the free coordinates in $U$ are constrained by the local perturbation
set, and the verifier is asked to certify the relevant extremal value of the
component output.

For verifier queries, it is convenient to work with the extremal raw
component value
\[
e_T(x,U) :=
\begin{cases}
\min_{z_T \in \mathcal{C}^{T}_{p,\epsilon}(x,U)} f_T(z_T), & \sigma(x)=1,\\
\max_{z_T \in \mathcal{C}^{T}_{p,\epsilon}(x,U)} f_T(z_T), & \sigma(x)=-1.
\end{cases}
\]
Then $\Delta_T(x,U)=\sigma(x)f_T(x_T)-\sigma(x)e_T(x,U)$. Suppose the verifier returns certified bounds
$\underline{e}_T(x,U) \leq e_T(x,U) \leq \overline{e}_T(x,U)$. These induce
deviation bounds
\[
    \underline{\Delta}_T(x,U) =
    \begin{cases}
    \max\{0,f_T(x_T)-\overline{e}_T(x,U)\}, & \sigma(x)=1,\\
    \max\{0,\underline{e}_T(x,U)-f_T(x_T)\}, & \sigma(x)=-1,
    \end{cases}
\]
and
\[
    \overline{\Delta}_T(x,U) =
    \begin{cases}
    \max\{0,f_T(x_T)-\underline{e}_T(x,U)\}, & \sigma(x)=1,\\
    \max\{0,\overline{e}_T(x,U)-f_T(x_T)\}, & \sigma(x)=-1.
    \end{cases}
\]
These bounds satisfy
\[
\underline{\Delta}_T(x,U) \leq \Delta_T(x,U) \leq \overline{\Delta}_T(x,U)
\qquad \text{for every } T \in \mathcal{H},\ U \subseteq T.
\]

Replacing $\Delta_T(x,U)$ in \eqref{eq:strong-milp} by
$\underline{\Delta}_T(x,U)$ gives the \emph{optimistic} MILP. Replacing it with
$\overline{\Delta}_T(x,U)$ gives the \emph{pessimistic} MILP.

To tighten these bounds, we bisect the current interval for the extremal value
$e_T(x,U)$. Given certified bounds
$\underline{e}_T(x,U) \leq e_T(x,U) \leq \overline{e}_T(x,U)$, let
$q_T(x,U)=(\underline{e}_T(x,U)+\overline{e}_T(x,U))/2$. If $\sigma(x)=1$, we
ask the verifier whether the minimum of $f_T$ over
$\mathcal{C}^{T}_{p,\epsilon}(x,U)$ is at least $q_T(x,U)$; if $\sigma(x)=-1$,
we ask whether the maximum is at most $q_T(x,U)$. The answer identifies which
half-interval still contains $e_T(x,U)$. Replacing
$[\underline{e}_T(x,U),\overline{e}_T(x,U)]$ by that surviving half and then
reapplying the formulas above yields refined bounds
$\underline{\Delta}_T(x,U)$ and $\overline{\Delta}_T(x,U)$.

\begin{lemma}
\label{lem:strong-bounds-milp}
Let $v^{\downarrow}$ and $v^{\uparrow}$ be the optimal values of the optimistic
and pessimistic MILPs, respectively, and let $v^*$ be the optimum of the exact
MILP. Then $v^{\downarrow} \leq v^* \leq v^{\uparrow}$.
Moreover, every optimal pessimistic solution is strongly sufficient, and if
$v^{\downarrow}=v^{\uparrow}$ then the corresponding pessimistic solution is a
cardinally-minimal strongly sufficient explanation.

\end{lemma}

\paragraph{Proof of Lemma~\ref{lem:strong-bounds-milp}.}
Let $\mathcal{F}(\widehat{\Delta})$ denote the feasible set of
\eqref{eq:strong-milp} when the last constraint uses the table
$\widehat{\Delta}_T(x,U)$. Since
$\underline{\Delta}_T(x,U) \leq \Delta_T(x,U) \leq \overline{\Delta}_T(x,U)$
for every $T \in \mathcal{H}$ and $U \subseteq T$, replacing the coefficients
in the last constraint by larger values can only shrink the feasible region.
Therefore,
\[
    \mathcal{F}(\overline{\Delta}) \subseteq \mathcal{F}(\Delta)
    \subseteq \mathcal{F}(\underline{\Delta}).
\]
Taking the minimum of $\sum_i b_i$ over these nested feasible sets yields
\[
    v^{\downarrow} \leq v^* \leq v^{\uparrow}.
\]

Now let $(b^{\uparrow},c^{\uparrow})$ be an optimal pessimistic solution and
set $S^{\uparrow} = \{ i \in [n] : b_i^{\uparrow}=1 \}$. Since
$(b^{\uparrow},c^{\uparrow}) \in \mathcal{F}(\overline{\Delta})
\subseteq \mathcal{F}(\Delta)$, it is feasible for the exact MILP. The first
three constraint families in \eqref{eq:strong-milp} force the unique active
case variable in each component $T$ to be $c_{T,U_T(S^{\uparrow})}=1$, so the
last exact constraint becomes
\[
    \sum_{T \in \mathcal{H}} \Delta_T\bigl(x,U_T(S^{\uparrow})\bigr)
    \leq |F(x)|.
\]
By Proposition~\ref{prop:strong-sufficiency-sum}, this is exactly the
condition for $S^{\uparrow}$ to be strongly sufficient.

Finally, if $v^{\downarrow}=v^{\uparrow}$, then the inequality above forces
$v^*=v^{\uparrow}$. Since $(b^{\uparrow},c^{\uparrow})$ is exact-feasible and
has objective value $v^{\uparrow}$, it is also optimal for the exact MILP.
Proposition~\ref{prop:strong-exact-milp} then implies that $S^{\uparrow}$ is a
cardinally-minimal strongly sufficient explanation. \hfill$\square$

\subsubsection{Mixed discrete/continuous variant of Algorithm~1}
\label{app:strong-mixed-algorithm}

The main text states Algorithm~1 only for continuous domains. When some
perturbation-aware local completion sets $\mathcal{C}^{T}_{p,\epsilon}(x,U)$ are finite, the
same optimistic/pessimistic MILP outer loop can exploit exact discrete
subproblems by replacing the local bounding step
with enumeration whenever $\mathcal{C}^{T}_{p,\epsilon}(x,U)$ is small enough. The resulting
mixed discrete/continuous variant is shown in
Algorithm~\ref{alg:strong-mixed-domain}.

\begin{algorithm}[H]
\caption{Strong-sufficiency algorithm for mixed discrete/continuous domains}
\label{alg:strong-mixed-domain}
\small
\begin{algorithmic}[1]
\Require Any-Order NAM $F$, input $x$, norm parameter $p \in [1,\infty]$, perturbation radius $\epsilon > 0$, signed label $\sigma(x)=2h_F(x)-1$, enumeration threshold $B_{\mathrm{enum}}$
\ForAll{$T \in \mathcal{H}$ and $U \subseteq T$ \textbf{in parallel}}
    \If{$U = \emptyset$}
        \State Set $\underline{\Delta}_T(x,U) \gets 0$ and $\overline{\Delta}_T(x,U) \gets 0$
    \ElsIf{$\mathcal{C}^{T}_{p,\epsilon}(x,U)$ is finite and $|\mathcal{C}^{T}_{p,\epsilon}(x,U)| \leq B_{\mathrm{enum}}$}
        \State Compute $\delta_T(x,U) \gets \sigma(x)f_T(x_T)-\min_{z_T \in \mathcal{C}^{T}_{p,\epsilon}(x,U)} \sigma(x)f_T(z_T)$ by enumeration
        \State Set $\underline{\Delta}_T(x,U) \gets \overline{\Delta}_T(x,U) \gets \delta_T(x,U)$
    \Else
        \State Compute sound bounds $\underline{e}_T(x,U) \leq e_T(x,U) \leq \overline{e}_T(x,U)$
        \State Convert $[\underline{e}_T(x,U),\overline{e}_T(x,U)]$ into $[\underline{\Delta}_T(x,U),\overline{\Delta}_T(x,U)]$
    \EndIf
\EndFor
\While{true}
    \State Solve \eqref{eq:strong-milp} with $\widehat{\Delta}_T(x,U)=\overline{\Delta}_T(x,U)$ for all $(T,U)$
    \State Let $S^{\uparrow} \gets \{i \in [n] : b_i^{\uparrow} = 1\}$ be the pessimistic explanation
    \State Solve \eqref{eq:strong-milp} with $\widehat{\Delta}_T(x,U)=\underline{\Delta}_T(x,U)$ for all $(T,U)$
    \State Let $S^{\downarrow} \gets \{i \in [n] : b_i^{\downarrow} = 1\}$ be the optimistic explanation
    \If{$|S^{\uparrow}| = |S^{\downarrow}|$}
        \State \Return $S^{\uparrow}$
    \EndIf
    \ForAll{$(T,U)$ with $\underline{\Delta}_T(x,U) < \overline{\Delta}_T(x,U)$ \textbf{in parallel}}
        \State Let $q_T(x,U) \gets (\underline{e}_T(x,U)+\overline{e}_T(x,U))/2$
        \State Use one verifier query on $\mathcal{C}^{T}_{p,\epsilon}(x,U)$ to determine which half-interval still contains $e_T(x,U)$
        \State Replace $[\underline{e}_T(x,U),\overline{e}_T(x,U)]$ by that surviving half
        \State Update $[\underline{\Delta}_T(x,U),\overline{\Delta}_T(x,U)]$ accordingly
    \EndFor
\EndWhile
\end{algorithmic}
\end{algorithm}

Whenever a local completion set is enumerated exactly, its deviation entry is
fixed from the outset and never refined again. In the fully discrete case, all
entries are exact after initialization, so the optimistic and pessimistic MILPs
coincide immediately. The same feasible-set nesting argument used in the proof
below yields correctness for this mixed-domain variant as well.

\paragraph{Proof of Theorem~\ref{thm:strong-convergence}.}
Throughout this proof, we assume that all activations in the network are
ReLUs. We begin by showing that the MILP template \eqref{eq:strong-milp} exactly
encodes strong sufficiency when supplied with the exact deviation table
$\Delta_T(x,U)$. Let $\mathcal{F}(\widehat{\Delta})$ denote the feasible set of
\eqref{eq:strong-milp} when the last constraint uses the table
$\widehat{\Delta}_T(x,U)$.

Take any feasible pair $(b,c) \in \mathcal{F}(\widehat{\Delta})$ and define
$S = \{ i \in [n] : b_i = 1 \}$. For every component $T \in \mathcal{H}$, the
constraint $\sum_{U \subseteq T} c_{T,U} = 1$ selects a unique subset
$U^*_T \subseteq T$ with $c_{T,U^*_T}=1$. The second family of constraints in
\eqref{eq:strong-milp} implies that $b_i = 0$ for every $i \in U^*_T$, while
the third family implies that $b_i = 1$ for every $i \in T \setminus U^*_T$.
Thus
\[
    U^*_T = \{ i \in T : b_i = 0 \} = T \cap \bar{S} = U_T(S).
\]
Conversely, for any set $S \subseteq [n]$, setting $b_i = 1$ iff $i \in S$ and
choosing exactly one active case variable per component via
$c_{T,U_T(S)} = 1$ and $c_{T,U}=0$ for all $U \neq U_T(S)$ satisfies the first
three families of constraints. Therefore, when $\widehat{\Delta}=\Delta$, the
last constraint in \eqref{eq:strong-milp} is precisely
\eqref{eq:strong-sufficiency-sum}, and the objective is exactly $|S|$. It
follows that the exact MILP computes a minimum-cardinality strongly sufficient
explanation.

Next, we use the interval bounds maintained by Algorithm~1. At initialization,
and after every refinement step, the bounds satisfy
\[
    \underline{\Delta}_T(x,U) \leq \Delta_T(x,U) \leq
    \overline{\Delta}_T(x,U)
    \qquad \text{for every } T \in \mathcal{H},\ U \subseteq T,
\]
because the local bounding routine always returns sound intervals and every
refinement step only shrinks such an interval while preserving the true optimum
inside it. Since larger coefficients in the last constraint of
\eqref{eq:strong-milp} only make feasibility harder, we obtain the nested
feasible regions
\[
    \mathcal{F}(\overline{\Delta}) \subseteq \mathcal{F}(\Delta)
    \subseteq \mathcal{F}(\underline{\Delta}).
\]
Let
\[
    v^{\uparrow} = \min\{\textstyle\sum_i b_i : (b,c) \in \mathcal{F}(\overline{\Delta})\},
    \qquad
    v^{\downarrow} = \min\{\textstyle\sum_i b_i : (b,c) \in \mathcal{F}(\underline{\Delta})\},
\]
and let $v^*$ be the optimum of the exact MILP $\mathcal{F}(\Delta)$. By the
set inclusion above,
\[
    v^{\downarrow} \leq v^* \leq v^{\uparrow}.
\]

Let $(b^{\uparrow},c^{\uparrow})$ be an optimal pessimistic solution and
$S^{\uparrow} = \{ i : b_i^{\uparrow}=1 \}$. Since
$(b^{\uparrow},c^{\uparrow}) \in \mathcal{F}(\overline{\Delta})
\subseteq \mathcal{F}(\Delta)$, the set $S^{\uparrow}$ is feasible for the exact
MILP and hence satisfies \eqref{eq:strong-sufficiency-sum}. By
Proposition~\ref{prop:strong-sufficiency-sum}, $S^{\uparrow}$ is strongly sufficient. Hence every pessimistic
iterate is strongly sufficient. Proposition~\ref{prop:strong-implies-sufficient} then implies that every such
iterate is also sufficient in the ordinary sense.

For optimality, suppose that the optimistic and pessimistic MILPs have
the same optimal objective value, i.e., $v^{\downarrow} = v^{\uparrow}$. The
inequality $v^{\downarrow} \leq v^* \leq v^{\uparrow}$ then forces
$v^* = v^{\uparrow}$. Since $S^{\uparrow}$ is an exact feasible solution with
objective value $v^{\uparrow}$, it attains the exact optimum and is therefore a
cardinally-minimal strongly sufficient explanation.

For termination, we apply the same interval-halving idea used in the
first-order NAM setting
\citep[App.~D.1, Proposition~5]{BaElLaSaKrAlKa26}, now to the finitely many
local deviation subproblems $(T,U)$. For each unresolved pair $(T,U)$, the
algorithm maintains an interval for the local extremal value $e_T(x,U)$. Because the
subproblem is finitely encoded and the local bounding routine returns rational
bounds, this interval has a finite binary precision budget. Each refinement
step bisects the current interval and thus removes one bit of uncertainty from
that local subproblem. Since Algorithm~1 refines all unresolved pairs in
parallel, the number of outer refinement rounds is bounded by the largest
precision budget of any one local subproblem, which is linear in the model
encoding size. After that many rounds, every remaining interval is exact, so
$\underline{\Delta}=\overline{\Delta}=\Delta$ and the two MILPs coincide with
the exact MILP. Therefore, Algorithm~1 must terminate.

For the per-round complexity statement, let
$r = \max_{T \in \mathcal{H}} |T|$. The algorithm must consider one deviation
entry for every nonempty free subset $U \subseteq T$, namely $2^{|T|}-1$
possibilities per component. Summing over all components gives at most
\[
    \sum_{T \in \mathcal{H}} (2^{|T|}-1) \leq |\mathcal{H}|\,2^r
\]
local deviation entries in a batched refinement pass. The MILP contains $n$
binary variables $b_i$ and one binary case variable $c_{T,U}$ for each pair
$(T,U)$, for a total of $n + \sum_{T \in \mathcal{H}} 2^{|T|}$ variables. The
first constraint family contributes $|\mathcal{H}|$ equalities; the second and
third together contribute
\[
    \sum_{T \in \mathcal{H}} \sum_{U \subseteq T} |T|
    \leq r \sum_{T \in \mathcal{H}} 2^{|T|}
\]
inequalities; and the deviation budget contributes one additional inequality.
Hence, for fixed $r$, each outer round performs two MILP solves and one batched
refinement pass, with polynomial-size MILPs and polynomially many local
deviation subproblems. Combined with the linear outer-round bound above, this
proves the claimed round complexity. \hfill$\square$

\subsection{Proofs for Section~\ref{sec:sparse-any-order}}

\paragraph{Proof of Lemma~\ref{lem:sparse-merged-components}.}
By definition, $C_1,\dots,C_m$ are the connected components of the
feature-interaction graph $\mathcal{G}_{\mathcal{H}}$, so they are pairwise
disjoint. Hence each input feature belongs to exactly one connected component.

Now fix any interaction scope $T \in \mathcal{H}$. For every pair of features
$i,j \in T$, the construction of $\mathcal{G}_{\mathcal{H}}$ places an edge
between $i$ and $j$. Thus the vertices of $T$ form a clique in
$\mathcal{G}_{\mathcal{H}}$, and in particular they all lie in the same
connected component. Therefore $T$ is contained in a unique merged component
$C_\ell$. \hfill$\square$

\paragraph{Proof of Theorem~\ref{thm:sparse-correctness}.}
Let
\[
    \widetilde{F}(x)=b+\sum_{\ell=1}^{m} G_\ell(x_{C_\ell})
\]
be the regrouped model from Section~\ref{sec:sparse-any-order}. By
construction, $\widetilde{F}(x)=F(x)$ for every input $x$. Moreover, by
Lemma~\ref{lem:sparse-merged-components}, every original interaction scope is
contained in a unique merged component, and distinct merged components share no
interactions.

Fix any candidate explanation $S \subseteq [n]$. Because the merged components
are disjoint, any choice of local completions for the components of
$\widetilde{F}$ can be concatenated into a unique global completion for $F$
that is consistent with the revealed coordinates $x_S$. Conversely, every
global completion for $F$ induces a compatible family of local completions for
$\widetilde{F}$. Therefore $S$ is sufficient for the original model $F$ if and
only if it is strongly sufficient for the regrouped model $\widetilde{F}$.

The assumptions of the theorem ensure that the local optimization problems for
the merged components of $\widetilde{F}$ satisfy the hypotheses of
Theorem~\ref{thm:strong-convergence}. Hence Algorithm~1, when run on
$\widetilde{F}$, terminates after a linear number of outer refinement rounds
and returns a minimum-cardinality strongly sufficient explanation
$S^{\star}$ for $\widetilde{F}$. By the equivalence above, $S^{\star}$ is a
sufficient explanation for the original model $F$.

To prove cardinal minimality for $F$, suppose toward contradiction that there
exists a sufficient explanation $S$ for $F$ with $|S| < |S^{\star}|$. By the
same equivalence, $S$ is strongly sufficient for $\widetilde{F}$. This
contradicts the minimum-cardinality guarantee returned by
Theorem~\ref{thm:strong-convergence}. Therefore $S^{\star}$ is a
cardinally-minimal sufficient explanation for $F$.

It remains to justify the complexity statement. The merged components can be
computed by a union-find pass over the interaction scopes: initialize one set
for each feature and, for every scope $T = \{i_1,\dots,i_q\}$, perform the
unions $\mathrm{union}(i_1,i_j)$ for $j=2,\dots,q$. Two features end up in the
same union-find class exactly when they are connected by a chain of overlapping
interaction scopes. This preprocessing step is linear up to inverse-Ackermann
factors in the total hyperedge-incidence size.

After preprocessing, every verifier or MILP subproblem in Algorithm~1 is posed
over a merged component $C_\ell$ of size at most
$k = \max_{\ell \in [m]} |C_\ell|$. Consequently, the per-round and
termination guarantees are exactly those of
Theorem~\ref{thm:strong-convergence}, with the worst-case exponential
dependence governed by $k$ rather than by the total number of features.
\hfill$\square$

\section{Additional Experimental Details}
\label{app:additional-experimental-details}

\subsection{Setup}

We evaluate both algorithmic settings introduced in the main text: the dual-MILP
algorithm for strong sufficiency (Section~\ref{sec:strong-sufficiency}) and its
sparse reduction with preprocessing (Section~\ref{sec:sparse-any-order}). We
compare both methods against two verifier-backed greedy baselines. The first is
a plain greedy baseline that starts from the full set of input features, visits
the features in a fixed canonical order, and after each tentative removal
checks sufficiency through a CROWN-style verifier ($\alpha$-CROWN in our
implementation) \citep{ZhWeChHsDa18}. The second is a
sensitivity-greedy baseline that uses the same remove-and-verify loop, but
orders features from least to most locally sensitive. The generic template
appears in Appendix~\ref{app:background} as
Algorithm~\ref{alg:greedy-baseline}; the two baselines differ only in the
choice of the ordering routine.

All experiments are run on a SLURM cluster. We launch one job per explained
test-set chunk, allocating a single GPU and 128~GB of RAM to each job. Unless noted
otherwise, we use $\alpha$-CROWN both to derive deviation bounds for our
methods and to answer the baselines' sufficiency queries, while Gurobi
\citep{Gu26} solves the optimistic and pessimistic MILP programs.

\subsection{Training and Explanation Hyperparameters}

Data
are split once into stratified 60/20/20 train/validation/test partitions with
seed 42. Continuous features are imputed with the median and standardized using
statistics fit on the training split only. Categorical features are imputed
with the most frequent value and ordinal-encoded, with unseen categories mapped
to $-1$ at evaluation time.

The default K-Order NAM training configuration uses ReLU units, hidden layer
widths $(64,64,32)$, dropout 0.1, feature dropout 0.05, learning rate $10^{-3}$,
weight decay $10^{-4}$, batch size 256, a maximum of 200 epochs, early stopping
with patience 20, and a \texttt{ReduceLROnPlateau} scheduler with patience 10
and multiplicative factor 0.5. The training script also records the exact model
order, component counts, maximum merged-group size, validation/test metrics,
and number of trained epochs in the saved metrics file for each run.

For the explanation experiments, the accompanying runner uses $\epsilon=0.1$ and
$p=\infty$ in the reported benchmark tables unless stated otherwise. The MILP
pipeline uses at most 10 outer tightening iterations by default, while the
greedy baselines use 20 $\alpha$-CROWN refinement steps per verification query.
The default discrete-domain threshold is 20 distinct values, the fallback
enumeration threshold for greedy verification is $10^6$, and the per-example
timeout is disabled by default. The sparse-setting experiments are obtained by
enabling the preprocessing flag in the same experiment runner.

\subsection{Reported Statistics}

The main-text tables report test accuracy, mean explanation size (as a
percentage of the total number of features), and mean wall-clock runtime over
the explained test set. To avoid reducing explanation-size behavior to a single
aggregate number, we also report the full empirical explanation-size
distribution through histograms/bar charts over explained examples. Runtime
variability is summarized with standard deviation information in the
corresponding plots, and the appendix ablation tables additionally report
medians where that helps characterize skewed runtime distributions. Because the
accompanying artifact stores the per-example outputs for every explained test
point, all reported means, medians, standard deviations, and histograms can be
recomputed directly from the saved result files.

\subsection{Compute Resources and External Requirements}

Training and explanation experiments are orchestrated through SLURM. The
accompanying experiment engine processes the test set in contiguous chunks and is
designed to launch one job per chunk; in our reported runs, each such job uses
a single GPU worker and 64~GB of RAM. The training code detects a CUDA GPU when
available and otherwise falls back to Apple MPS or CPU execution, while the
explanation code records per-example wall-clock time in the saved results so
that total benchmark runtime can be reconstructed exactly from the artifact.

The MILP-based experiments require a Gurobi installation and a valid Gurobi
license. Academic users can request a free academic license. Beyond Gurobi, the
remaining external assets used in the paper---including the public tabular
datasets and the open-source software dependencies used by the training and
verification pipelines---are available under their standard published terms and
do not require an additional commercial license.

\subsection{Dataset and Model Summary}

We use nine public tabular benchmarks: Adult, Bike Sharing, Breast Cancer
Wisconsin, California Housing, COMPAS, Diabetes, German Credit, Heart Disease,
and HELOC. Adult, Bike Sharing, COMPAS, German Credit, and Heart Disease are
mixed discrete--continuous tasks, while Breast Cancer, California Housing,
Diabetes, and HELOC are fully continuous. The table below reports the exact
split sizes and feature-type counts for all benchmarks used in the main-text
experiments.

For every benchmark, we evaluate explanations on the full held-out test split,
so the number of explained examples in
Section~\ref{sec:empirical-evaluation} coincides with the test-set
size listed below.

\begin{center}
\scriptsize
\setlength{\tabcolsep}{4pt}
\resizebox{\linewidth}{!}{%
\begin{tabular}{lrrrrrrr}
\toprule
Dataset & Total & Train & Val & Test & Features & Continuous & Categorical \\
\midrule
Adult & 48{,}842 & 29{,}305 & 9{,}768 & 9{,}769 & 14 & 6 & 8 \\
Bike Sharing & 17{,}379 & 10{,}427 & 3{,}476 & 3{,}476 & 12 & 4 & 8 \\
Breast Cancer & 569 & 341 & 114 & 114 & 30 & 30 & 0 \\
California Housing & 20{,}640 & 12{,}384 & 4{,}128 & 4{,}128 & 8 & 8 & 0 \\
COMPAS & 5{,}278 & 3{,}166 & 1{,}056 & 1{,}056 & 13 & 2 & 11 \\
Diabetes & 768 & 460 & 154 & 154 & 8 & 8 & 0 \\
German Credit & 1{,}000 & 600 & 200 & 200 & 20 & 7 & 13 \\
Heart Disease & 303 & 181 & 61 & 61 & 13 & 5 & 8 \\
HELOC & 10{,}000 & 6{,}000 & 2{,}000 & 2{,}000 & 22 & 22 & 0 \\
\bottomrule
\end{tabular}}
\end{center}

\subsection{Additional Strong-Sufficiency Curves}
\label{app:additional-strong-sufficiency-curves}

The following figures show the strong-sufficiency plots omitted from the main text.

\begin{figure}[H]
\centering
\makebox[\linewidth][c]{\includegraphics[width=1.05\linewidth]{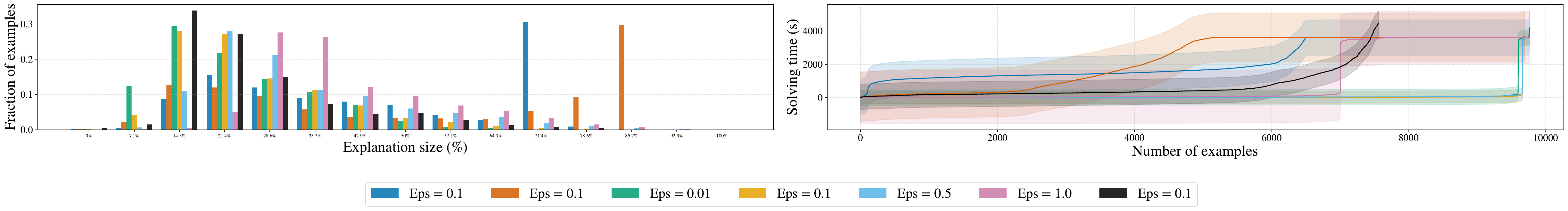}}
\caption{Adult: combined explanation-size and runtime plot under strong sufficiency.}
\label{fig:strong-sufficiency-curves-adult}
\end{figure}

\begin{figure}[H]
\centering
\includegraphics[width=0.97\linewidth]{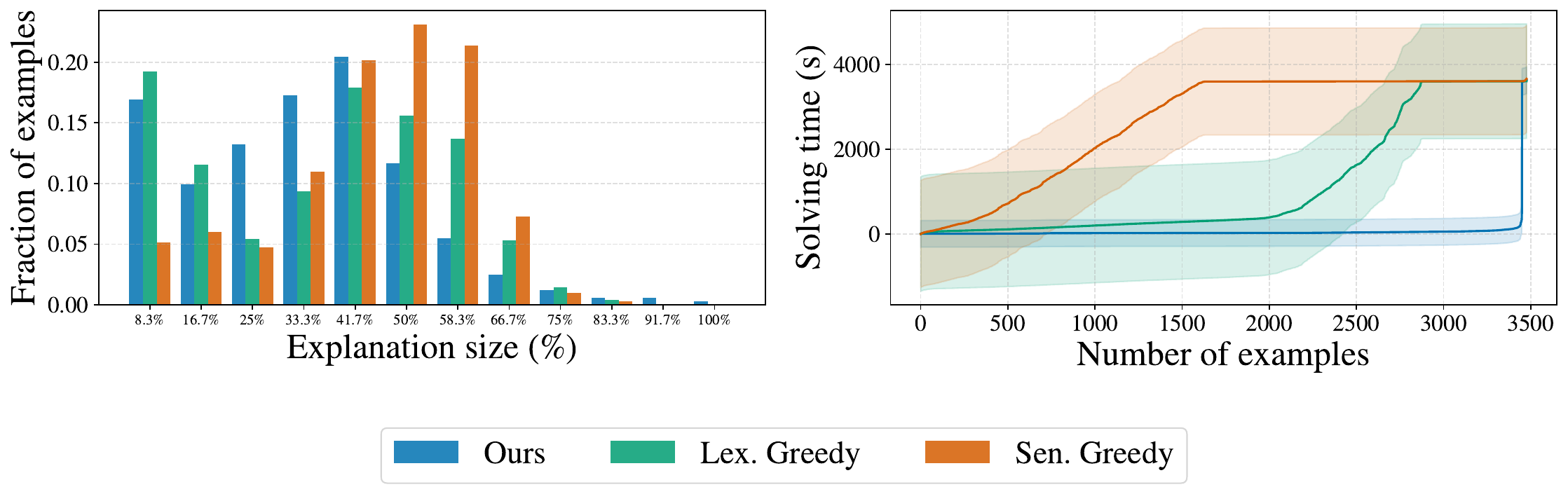}
\caption{Bike Sharing: combined explanation-size and runtime plot under strong sufficiency.}
\label{fig:strong-sufficiency-curves-bike-sharing}
\end{figure}

\begin{figure}[H]
\centering
\includegraphics[width=0.97\linewidth]{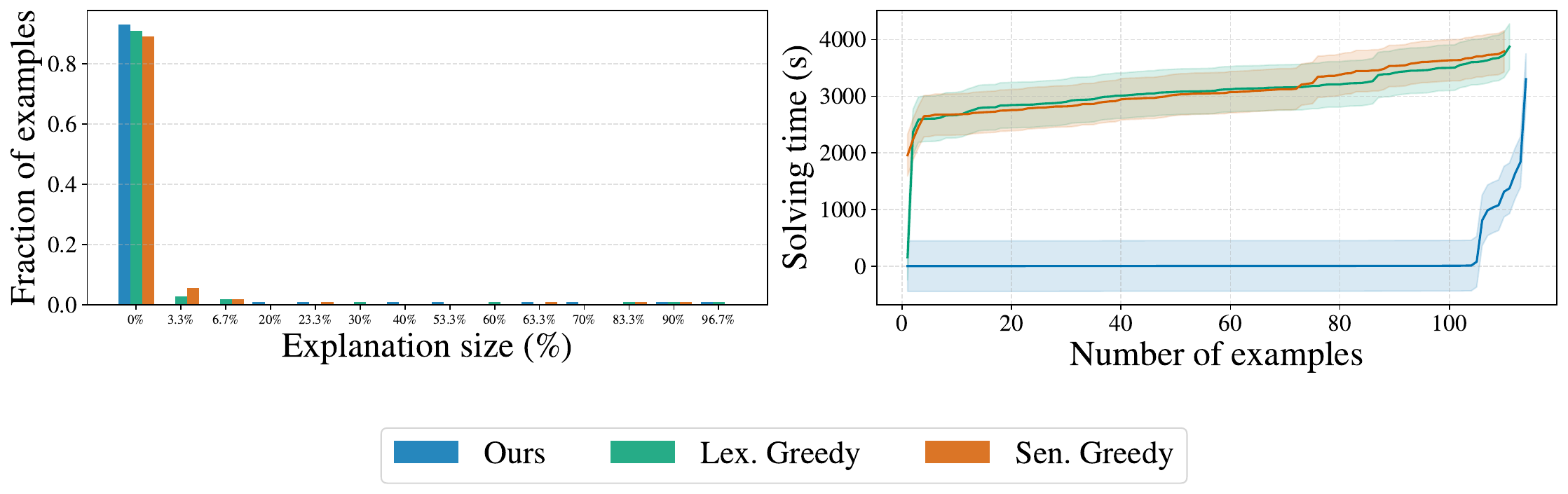}
\caption{Breast Cancer: combined explanation-size and runtime plot under strong sufficiency.}
\label{fig:strong-sufficiency-curves-breast-cancer}
\end{figure}

\begin{figure}[H]
\centering
\includegraphics[width=0.97\linewidth]{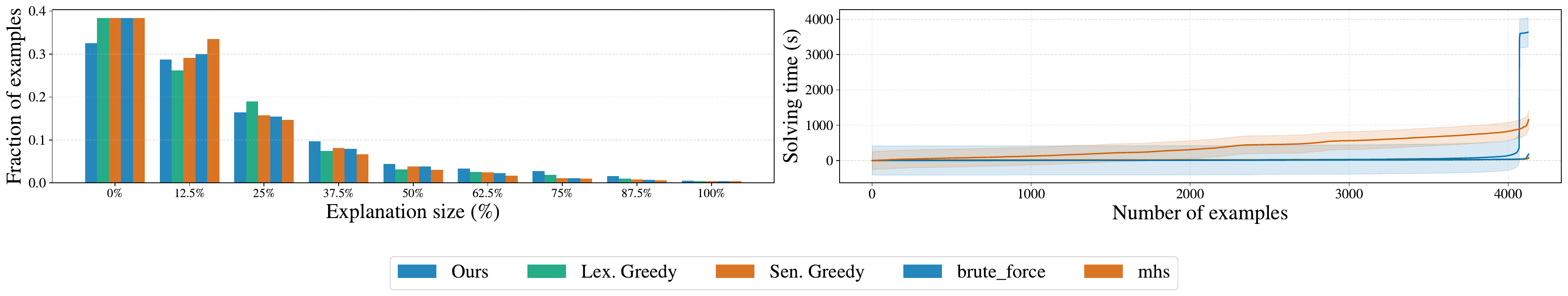}
\caption{California Housing: combined explanation-size and runtime plot under strong sufficiency.}
\label{fig:strong-sufficiency-curves-california-housing}
\end{figure}

\begin{figure}[H]
\centering
\includegraphics[width=0.97\linewidth]{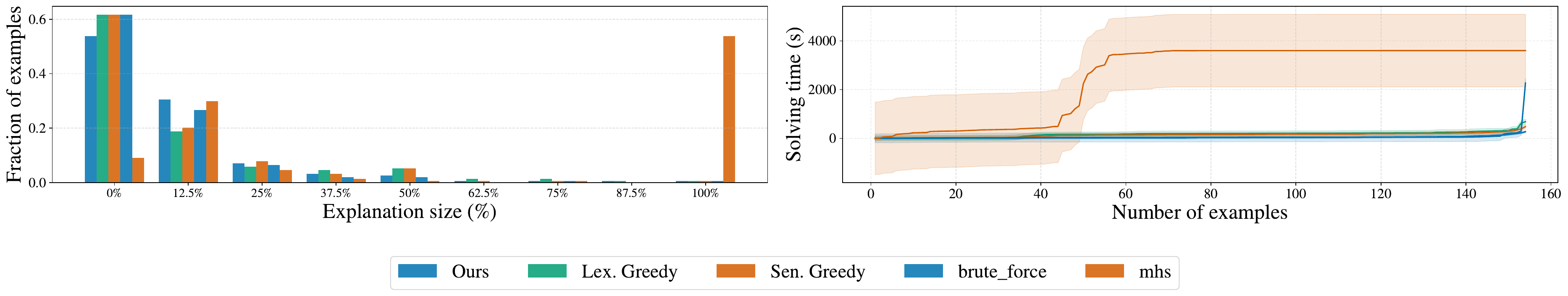}
\caption{Diabetes: combined explanation-size and runtime plot under strong sufficiency.}
\label{fig:strong-sufficiency-curves-diabetes}
\end{figure}

\begin{figure}[H]
\centering
\includegraphics[width=0.97\linewidth]{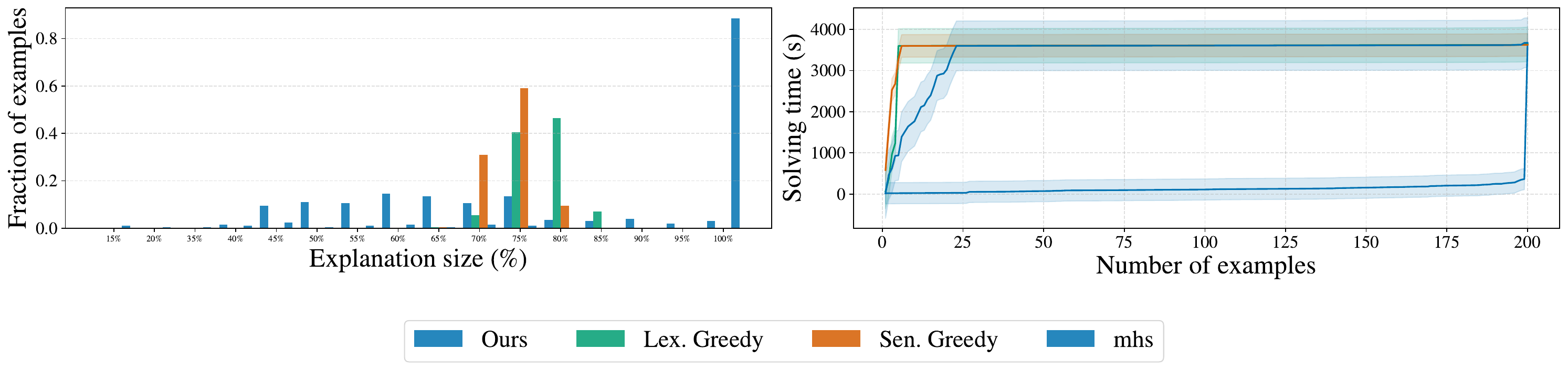}
\caption{German Credit: combined explanation-size and runtime plot under strong sufficiency.}
\label{fig:strong-sufficiency-curves-german-credit}
\end{figure}

\begin{figure}[H]
\centering
\includegraphics[width=0.97\linewidth]{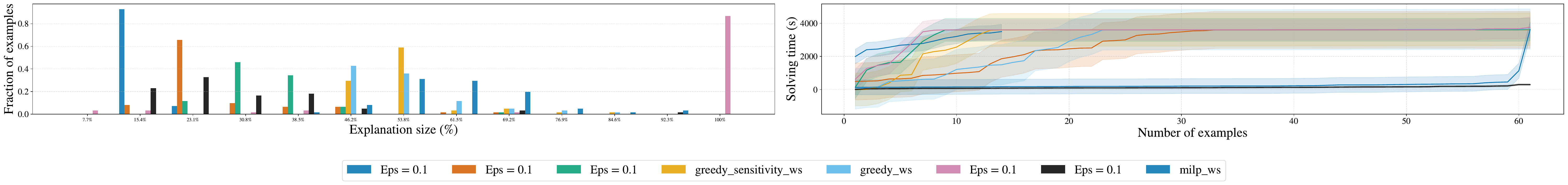}
\caption{Heart Disease: combined explanation-size and runtime plot under strong sufficiency.}
\label{fig:strong-sufficiency-curves-heart-disease}
\end{figure}

\begin{figure}[H]
\centering
\includegraphics[width=0.97\linewidth]{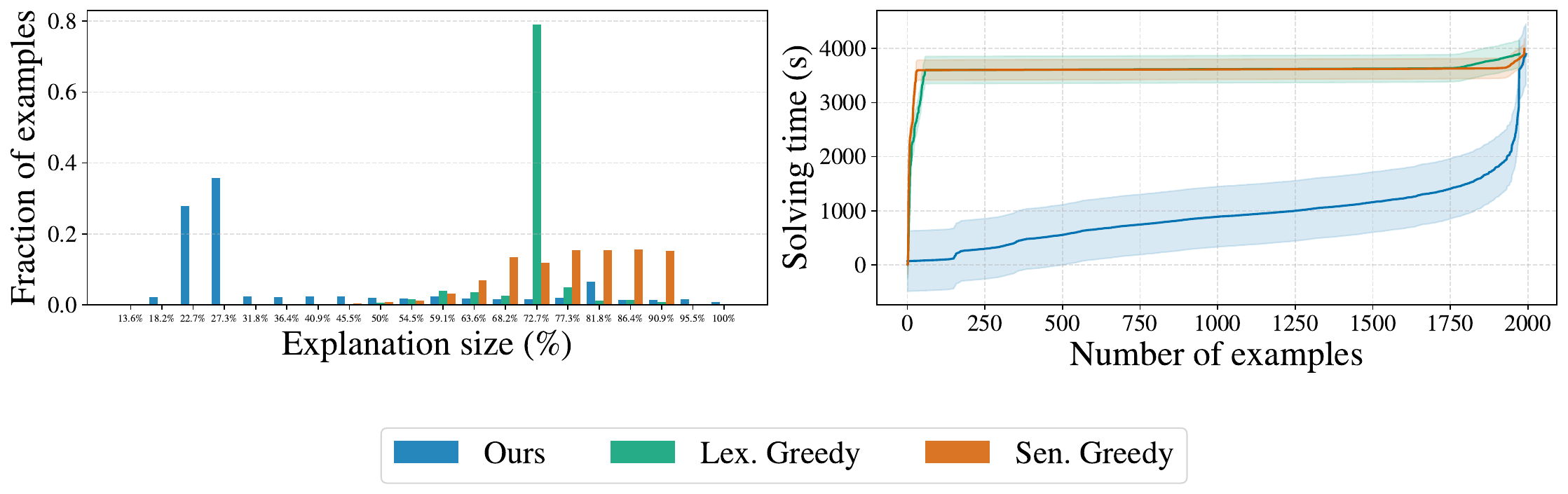}
\caption{HELOC: combined explanation-size and runtime plot under strong sufficiency.}
\label{fig:strong-sufficiency-curves-heloc}
\end{figure}

\section{Ablation Studies}
\label{app:ablation-studies}

\subsection{Sensitivity to \texorpdfstring{$\epsilon$}{epsilon}}

As $\epsilon$ grows, certificates generally get larger, and runtime tails become
heavier, though most inputs remain easy and a small set of hard examples drives
the worst runtimes.

\begin{figure}[H]
\centering
\includegraphics[width=0.97\linewidth]{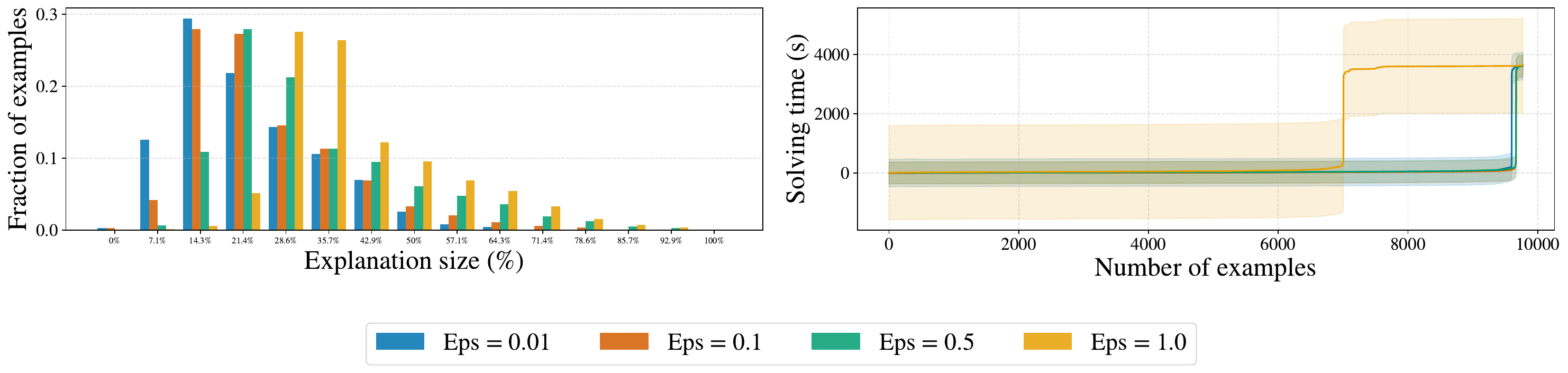}
\caption{Adult under varying $\epsilon$. Combined explanation-size and runtime plot.}
\label{fig:adult-epsilon-sweep}
\end{figure}

\subsection{Sensitivity to the Admissible Input Space}
\label{app:whole-space}

The $\epsilon$-ball is a local special case of the more general distance-restricted
sufficiency setting: on a bounded domain, a sufficiently large radius covers
the entire admissible input space. To demonstrate that our method applies
without modification in this unrestricted regime, we re-run the Heart
Disease benchmark (strong-sufficiency setting) using the full
training-data range $[\min_i x_i, \max_i x_i]$ for each continuous feature
as its free region, instead of an $\epsilon$-ball.

\begin{table}[H]
\centering
\scriptsize
\setlength{\tabcolsep}{6pt}
\caption{Heart Disease under strong sufficiency: $\epsilon=0.1$ vs. the full admissible input space. Each cell reports mean size $\pm$ sd (\%) / mean runtime $\pm$ sd (s) / solved instances.}
\label{tab:whole-space}
\resizebox{\linewidth}{!}{%
\begin{tabular}{lccc}
\toprule
Setting & Ours & Greedy & Greedy Sensitivity \\
\midrule
$\epsilon=0.1$ & 29.1$\pm$14.4\% / 130.5$\pm$54.8 / 61/61 & 27.1$\pm$10.3\% / 2765.5$\pm$1121.9 / 48/61 & 34.2$\pm$7.5\% / 3369.3$\pm$695.3 / 11/61 \\
Whole admissible space & 61.2$\pm$10.5\% / 303.0$\pm$453.1 / 60/61 & 53.5$\pm$8.8\% / 2820.5$\pm$1223.1 / 51/61 & 53.5$\pm$7.5\% / 3196.4$\pm$996.5 / 13/61 \\
\bottomrule
\end{tabular}}
\end{table}

As expected, every method's certified explanations grow once the
perturbation space is unbounded, since more features must be fixed to
guarantee the prediction cannot change anywhere in the admissible space.
Our method continues to solve nearly every instance (60/61) and remains
the fastest of the three; the sensitivity-guided greedy baseline is hit
hardest by the harder setting, solving only 13/61 instances within budget.
Removing the distance restriction makes the problem harder for every
method, but our relative advantage persists.

\subsection{First- vs. Second-Order NAMs}

To isolate the effect of pairwise interactions, we compare three model classes
on four representative datasets: a first-order NAM baseline, a full
second-order NAM solved with the strongly-sufficient algorithm, and a
sparse second-order NAM solved with the sparse algorithm.
Adding second-order interactions improves predictive accuracy on three of
the four datasets (Bike Sharing, California Housing, COMPAS); on Heart
Disease, all three models tie exactly. The explanation-size picture now
favors simpler models: the full second-order NAM yields the largest mean
certificate on every dataset here, consistent with
Section~\ref{sec:empirical-evaluation}'s observation that strong
sufficiency carries no size guarantee relative to a simpler model. The
sparse second-order variant recovers the smallest mean certificate on
COMPAS and Heart Disease, while the first-order NAM is smallest on Bike
Sharing and narrowly ahead of sparse on California Housing. Runtime still
reflects the added expressive power: the first-order NAM is consistently
fastest, while the sparse second-order pipeline incurs the largest overhead
on Bike Sharing, COMPAS, and Heart Disease.

\vspace{0.4em}
\begin{table}[H]
\centering
\scriptsize
\setlength{\tabcolsep}{3.5pt}
\caption{First- vs. second-order NAM ablation on four datasets at $\epsilon=0.1$ on the full test split.}
\label{tab:first-vs-second-order-ablation}
\resizebox{\linewidth}{!}{%
\begin{tabular}{llccccc}
\toprule
Dataset & Model & Acc. & Size mean & Size med & Time mean (s) & Time med (s) \\
\midrule
Bike Sharing & First-order & 0.8800 & 1.74 & 1.0 & 1.651 & 1.082 \\
Bike Sharing & Second-order & 0.9232 & 4.02 & 4.0 & 61.279 & 26.505 \\
Bike Sharing & Sparse second-order & 0.9163 & 2.82 & 3.0 & 242.984 & 118.560 \\
\midrule
California Housing & First-order & 0.8619 & 1.22 & 1.0 & 5.786 & 1.767 \\
California Housing & Second-order & 0.8886 & 1.57 & 1.0 & 71.860 & 7.564 \\
California Housing & Sparse second-order & 0.8777 & 1.26 & 1.0 & 49.813 & 18.113 \\
\midrule
COMPAS & First-order & 0.6648 & 3.69 & 4.0 & 3.906 & 1.138 \\
COMPAS & Second-order & 0.6742 & 6.58 & 6.0 & 26.411 & 18.285 \\
COMPAS & Sparse second-order & 0.6799 & 2.56 & 2.0 & 40.424 & 30.043 \\
\midrule
Heart Disease & First-order & 0.8525 & 2.85 & 3.0 & 1.069 & 1.035 \\
Heart Disease & Second-order & 0.8525 & 3.79 & 3.0 & 130.520 & 116.814 \\
Heart Disease & Sparse second-order & 0.8525 & 2.46 & 2.0 & 182.618 & 98.464 \\
\bottomrule
\end{tabular}}
\end{table}

\begin{figure}[H]
\centering
\includegraphics[width=0.97\linewidth]{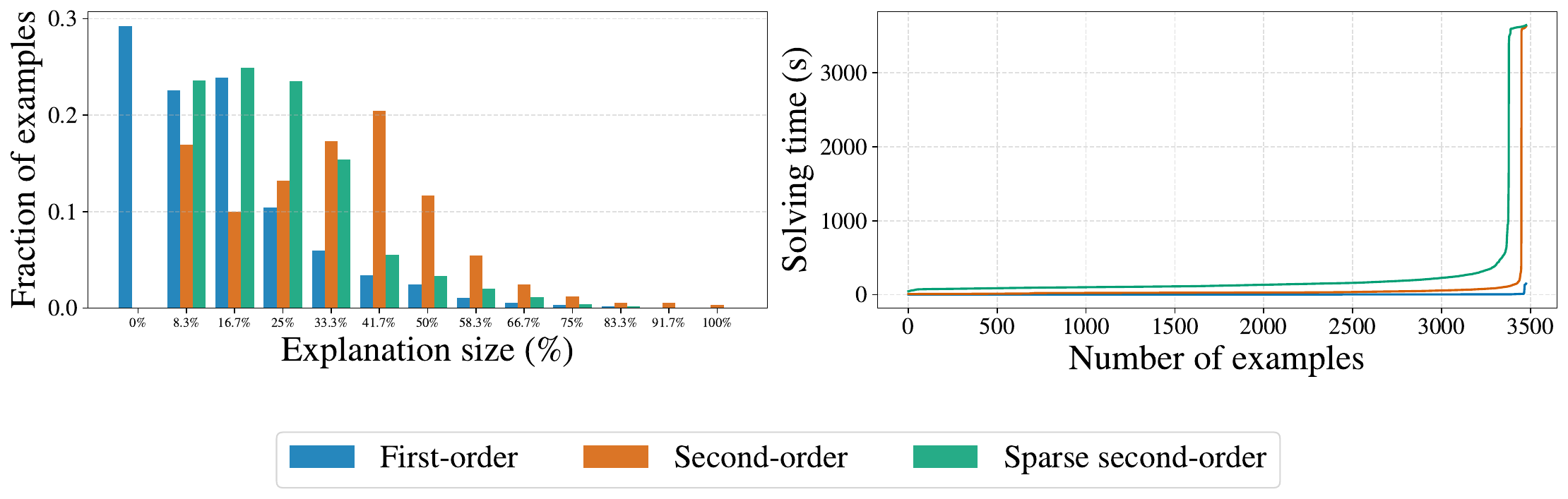}
\caption{Bike Sharing: combined explanation-size and runtime plots for the first-vs-second-order ablation across explained test points.}
\label{fig:first-vs-second-order-ablation-bike}
\end{figure}

\begin{figure}[H]
\centering
\includegraphics[width=0.97\linewidth]{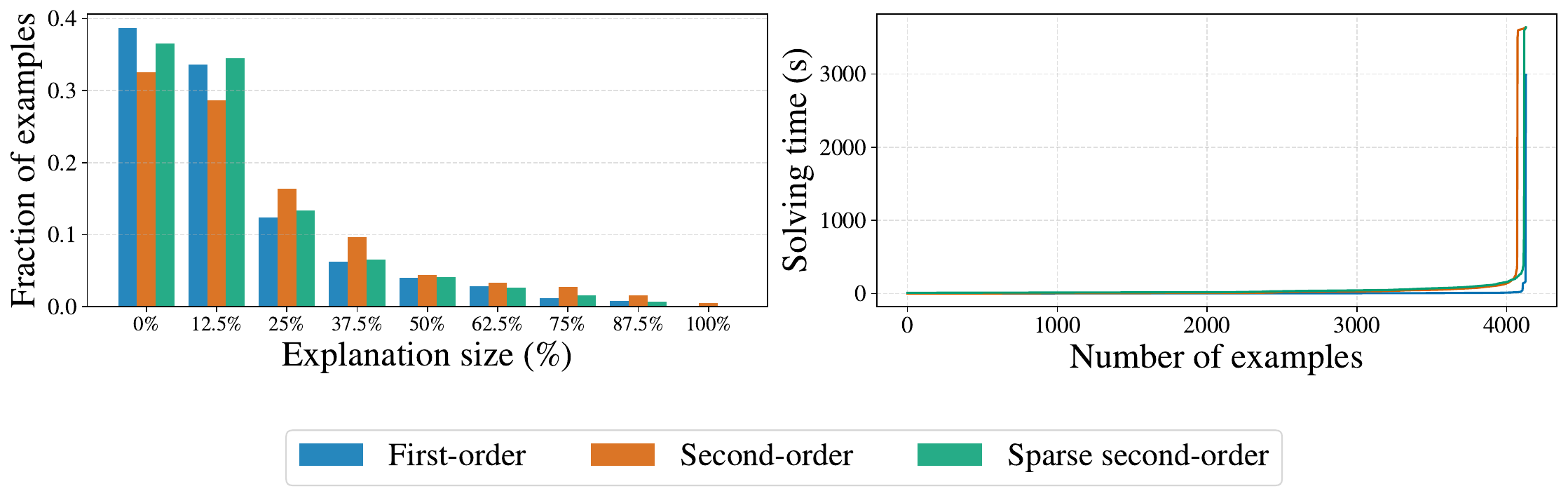}
\caption{California Housing: combined explanation-size and runtime plots for the first-vs-second-order ablation across explained test points.}
\label{fig:first-vs-second-order-ablation-california}
\end{figure}

\begin{figure}[H]
\centering
\includegraphics[width=0.97\linewidth]{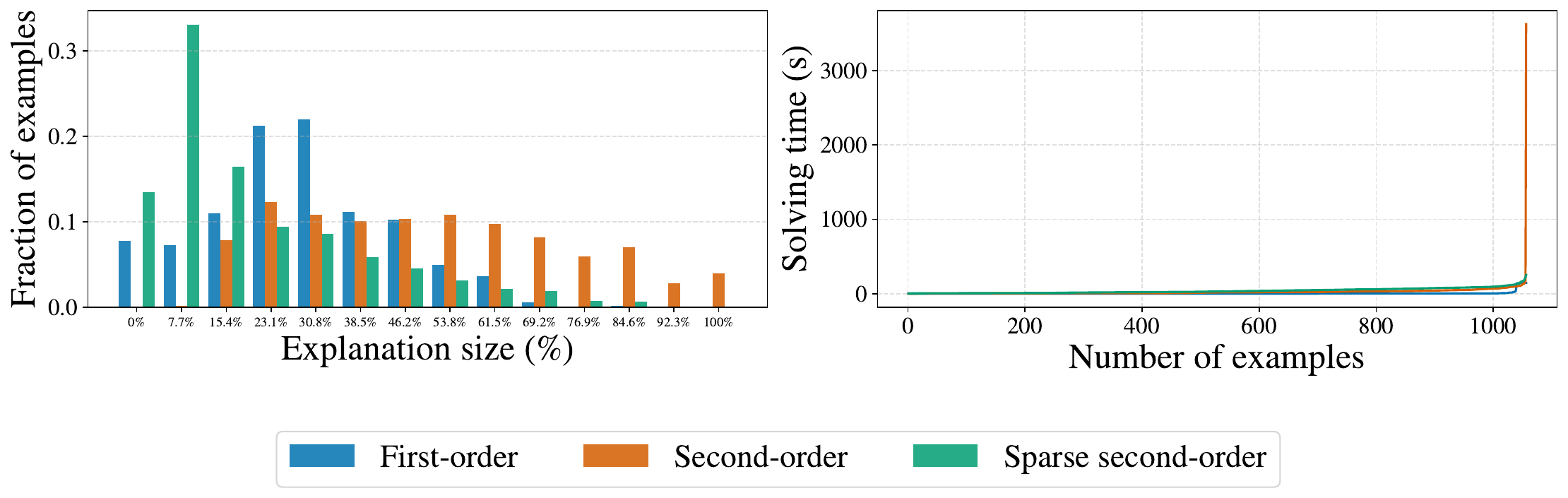}
\caption{COMPAS: combined explanation-size and runtime plots for the first-vs-second-order ablation across explained test points.}
\label{fig:first-vs-second-order-ablation-compas}
\end{figure}

\begin{figure}[H]
\centering
\includegraphics[width=0.97\linewidth]{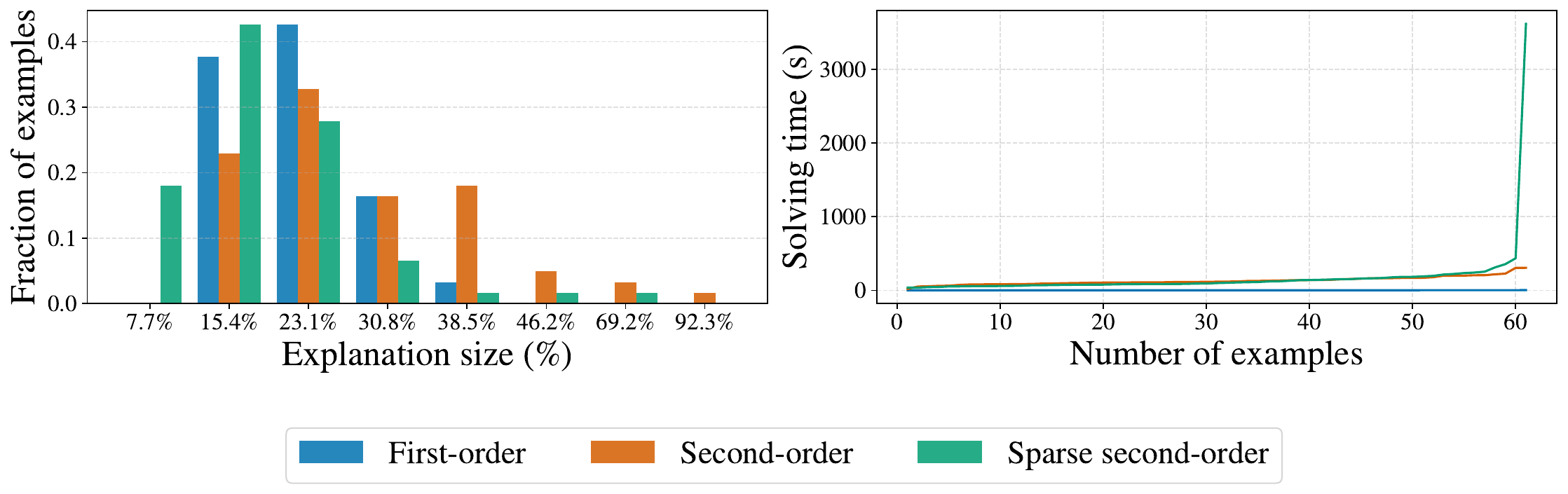}
\caption{Heart Disease: combined explanation-size and runtime plots for the first-vs-second-order ablation across explained test points.}
\label{fig:first-vs-second-order-ablation-heart}
\end{figure}

Figures~\ref{fig:first-vs-second-order-ablation-bike}--\ref{fig:first-vs-second-order-ablation-heart}
show the corresponding per-example distributions across the full test split
of each dataset.

\subsection{Preliminary Analysis: When Does the Gap Over Greedy Grow?}
\label{app:margin-correlation}

Section~\ref{sec:empirical-evaluation} observes that our method's advantage
over the greedy baselines varies substantially across benchmarks. As a first
step toward quantifying this, we measure, for each example solved by both
our method and canonical-order greedy under strong sufficiency, the
Pearson correlation between the gap (greedy's explanation size minus ours)
and the margin $|F(x)|$ at that example, pooling and per-dataset.

\begin{table}[H]
\centering
\scriptsize
\setlength{\tabcolsep}{8pt}
\caption{Correlation between the ours-vs-greedy size gap and $|F(x)|$, under strong sufficiency ($\epsilon=0.1$).}
\label{tab:margin-correlation}
\begin{tabular}{lcc}
\toprule
Dataset & $N$ & $r$(gap, $|F(x)|$) \\
\midrule
Adult & 6477 & 0.233 \\
Bike Sharing & 3017 & 0.057 \\
Breast Cancer & 105 & $-$0.138 \\
California Housing & 4070 & 0.360 \\
COMPAS & 357 & 0.787 \\
Diabetes & 154 & $-$0.055 \\
Heart Disease & 48 & 0.648 \\
HELOC & 70 & $-$0.255 \\
\midrule
Pooled & 14302 & 0.158 \\
\bottomrule
\end{tabular}
\end{table}

German Credit is omitted: canonical greedy times out on all but 4 of its
200 instances, too few for a meaningful correlation. Among the rest, the
pooled correlation is positive and statistically significant given the
sample size, but modest in magnitude and highly heterogeneous across
datasets, strong on COMPAS and Heart Disease but flat or slightly negative
on Breast Cancer, Diabetes, and HELOC. This is consistent with, though more
mixed than, the intuition that our method's advantage grows with decision
margin; it does not yet account for the complementary factor, the spread
of explanation sizes across alternative sufficient subsets (which requires
repeated greedy runs under randomized feature orderings), left for future
work.

\subsection{Higher-Order NAMs}

We also study how increasing interaction order affects the sparse family on
Bike Sharing. In this controlled study,
accuracy improves monotonically with order, from 0.9163 at order 2 to 0.9180
at order 3 and 0.9295 at order 4. The improved predictive performance comes at
a cost: the mean explanation size increases from 2.82 at order 2 to 2.90 at
order 3, and the mean solve time rises from 242.984s to 913.630s over the
same range. At order 4, the largest merged components grow further and
convergence becomes substantially harder: only 1096 of 3450 examples
(31.8\%) produce a usable explanation within the per-example time budget
(3600s), compared to essentially full coverage at orders 2 and 3; the
statistics below for order 4 are computed over that smaller, converged
subset only; and are likely biased toward comparatively easier instances. Thus, higher-order interactions continue to improve accuracy, but
certified explanations become both larger and substantially more expensive
to compute, to the point of a practical tractability wall at order 4 on
this benchmark.

\vspace{0.4em}
\begin{table}[H]
\centering
\scriptsize
\setlength{\tabcolsep}{6pt}
\caption{Bike Sharing higher-order ablation for sparse models at $\epsilon=0.1$ (orders 2--4; full test split). Order 4's statistics are computed only over the 1096/3450 (31.8\%) examples that produced a usable explanation within the 3600s per-example budget; see text.}
\label{tab:higher-order-ablation}
\begin{tabular}{cccccc}
\toprule
Order & Acc. & Size mean & Size med & Time mean (s) & Time med (s) \\
\midrule
2 & 0.9163 & 2.82 & 3.0 & 242.984 & 118.560 \\
3 & 0.9180 & 2.90 & 3.0 & 913.630 & 634.051 \\
4 & 0.9295 & 3.37 & 3.0 & 3500.879 & 3735.853 \\
\bottomrule
\end{tabular}
\end{table}

\subsection{Comparison Against Exact and Subset-Minimal Baselines}
\label{app:exact-baselines}

The greedy baselines compared against in the main text are, by construction,
only subset-minimal, which could in principle account for some of our
method's advantage. We therefore compare against two further baselines
spanning the subset-minimal and cardinally-minimal ends of the literature,
under the strong-sufficiency setting (Section~\ref{sec:strong-sufficiency}),
holding the verifier fixed across all methods. Any Any-Order NAM can
equivalently be written as a standard neural network with sparse,
block-structured weight matrices, so general-purpose explanation algorithms
apply to it directly without modification; both baselines here, and the two
greedy baselines in the main text, treat the model exactly this way, so any
gains we observe reflect how well a method exploits the additive
interaction structure rather than a stronger verifier or a weaker
certification standard.

\textbf{VeriX+.} VeriX~\citep{WuWuBa23} is, with the verifier held fixed, a
sensitivity-guided sequential remove-and-verify procedure, coinciding with
our sensitivity-guided greedy baseline. We instead compare against VeriX+
\citep{WuLiWuBa26}, which is algorithmically distinct: a QuickXplain-based
divide-and-conquer search~\citep{Ju04} that confirms a whole block of
features as safe to free with a single query whenever possible, using
far fewer oracle queries than one-at-a-time removal. On the full Adult test
set, our method solves every instance (9665/9769, mean size
25.7$\pm$13.2\%, median runtime 18.3s), whereas VeriX+ solves 7562/9769
within the same per-job time budget, with mean size 25.5$\pm$13.8\% and
median runtime 332.1s on the instances it solves. Explanation size is
essentially tied between the two methods here; the runtime advantage of our
method remains an order of magnitude, and it additionally solves every
instance within budget, whereas VeriX+ does not.

\textbf{Brute-force and MHS.} We also compare against two baselines that
target cardinal minimality directly: an exact brute-force search over
subsets of increasing cardinality, and a minimum-hitting-set (MHS)
enumeration algorithm exploiting the duality between contrastive and
sufficient explanations~\citep{IgNaMa19,BaElGu25}, solved with the
MaxSAT solver RC2~\citep{IgMoMa19}. Both are restricted to lower-dimensional
benchmarks, since exact search is not tractable at higher dimensionality.

\begin{table}[H]
\centering
\scriptsize
\setlength{\tabcolsep}{4pt}
\caption{Exact cardinally-minimal baselines vs. our method under strong sufficiency ($\epsilon=0.1$). Each cell reports mean size $\pm$ sd / median runtime (s) / solved instances.}
\label{tab:exact-baselines}
\begin{tabular}{lccc}
\toprule
Dataset (\# features) & Ours & Brute force & MHS \\
\midrule
Diabetes (8) & 0.82$\pm$1.33 / 29.0 / 154/154 & 0.62$\pm$1.13 / 41.0 / 154/154 & 4.81$\pm$3.52 / 3600.1 / 72/154 \\
California Housing (8) & 1.57$\pm$1.74 / 7.6 / 4070/4125 & 1.26$\pm$1.49 / 6.5 / 4128/4128 & 1.17$\pm$1.41 / 324.9 / 4128/4128 \\
German Credit (20) & 12.99$\pm$2.85 / 110.3 / 199/200 & n/a / n/a / 0/200 & 18.85$\pm$3.40 / 3607.2 / 23/200 \\
Heart Disease (13) & 3.79$\pm$1.87 / 116.8 / 61/61 & 2.07$\pm$0.27 / 2783.6 / 14/61 & 11.77$\pm$3.32 / 3603.1 / 8/61 \\
\bottomrule
\end{tabular}
\end{table}

Brute-force only completes within budget on the two lowest-dimensional
benchmarks (Diabetes, California Housing with 8 features each); on those,
it is close to our method's size, as expected since it recovers the true
cardinally-minimal ordinary-sufficient explanation. It times out entirely
on German Credit's 20 features. MHS solves a minority of instances
everywhere except California Housing, and takes orders of magnitude longer
than our method where it does solve. Taken together, exact cardinally-minimal
search and MHS do not scale to the dimensionalities in our main
benchmarks, which is precisely why the main text compares against
verifier-backed greedy baselines instead; our method remains close to the
true optimum where one can be computed at all.

\subsection{Validating the Sparsity Assumption via SIAN}
\label{app:sian}

The sparsity assumption of Section~\ref{sec:sparse-any-order} is controlled
by construction in our main experiments, which raises the question of
whether small merged components ($s$ small) also arise when interactions
are discovered from data rather than imposed by us. We investigate this
using Sparse Interaction Additive Networks (SIAN)~\citep{EnLi22}: a base DNN
is trained, its most important interactions are detected via hierarchical
Archipelago-based scoring, and a sparse Any-Order NAM is built from exactly
those detected interactions. We apply this pipeline to the Diabetes and
Heart Disease benchmarks and then run our exact sparse-setting algorithm on
the resulting models.

\begin{table}[H]
\centering
\scriptsize
\setlength{\tabcolsep}{6pt}
\caption{SIAN-detected interaction models: predictive accuracy and certified explanations under strong sufficiency (full test split, $\epsilon=0.1$).}
\label{tab:sian}
\begin{tabular}{lccccc}
\toprule
Dataset & DNN acc. & SIAN acc. & Order & $s$ & Size mean $\pm$ sd (\%) / Runtime mean $\pm$ sd (s) \\
\midrule
Diabetes & 0.7532 & 0.7273 & 4 & 4 & 8.4$\pm$15.0 / 66.1$\pm$63.7 \\
Heart Disease & 0.8689 & 0.8361 & 3 & 3 & 16.5$\pm$12.8 / 59.3$\pm$40.6 \\
\bottomrule
\end{tabular}
\end{table}

Both benchmarks yield data-discovered merged components of size 3--4 out of
8 and 13 features respectively, i.e., exactly the small-$s$ regime our
method targets, with no component size imposed by us. The SIAN models
remain competitive with the underlying DNN in accuracy, and our exact
method solves every instance (100\% converged, 0 timeouts on both
benchmarks), supporting the practical relevance of the sparsity assumption
on models derived from standard DNNs rather than constructed by hand.

\subsection{A Qualitative Case Study}
\label{app:case-study}

To illustrate what a certified explanation looks like in practice, we walk
through test patient 38 of the Heart Disease benchmark (strong-sufficiency
model of Table~\ref{tab:strong-sufficiency-results}). The model predicts no
heart disease for this patient. Our method returns the five-feature
certificate $\{\text{cp}, \text{restecg}, \text{exang}, \text{slope},
\text{thal}\}$ (chest-pain type, resting ECG, exercise-induced
angina, ST-segment slope, and thalassemia status), and because our MILP
converged, this additionally certifies that no explanation with four or
fewer features exists. The sensitivity-guided greedy baseline instead
returns an eight-feature set, $\{\text{sex}, \text{cp}, \text{restecg},
\text{thalach}, \text{exang}, \text{oldpeak}, \text{slope}, \text{ca}\}$,
which is also a valid sufficient explanation but nearly twice as large,
presenting a clinician with more factors than are actually needed to
certify the prediction.

\end{document}